\documentclass[acmtog]{acmart}

\usepackage{booktabs} 

\usepackage[ruled]{algorithm2e} 

\SetAlFnt{\small}
\SetAlCapFnt{\small}
\SetAlCapNameFnt{\small}
\SetAlCapHSkip{0pt}
\IncMargin{-\parindent}

\usepackage{makecell}
\usepackage{epsfig}
\usepackage{CJKutf8} 
\usepackage{booktabs} 
\usepackage{stmaryrd} 
\usepackage{multirow} 
\usepackage{soul} 
\usepackage{subcaption} 
\usepackage{color}
\definecolor{darkyellow}{RGB}{100,100,32}

\setcitestyle{square}

\begin{document}
\title{DesnowNet: Context-Aware Deep Network for Snow Removal}

\author{Yun-Fu Liu}
\affiliation{%
  \institution{Viscovery}
}
\email{ yunfuliu@gmail.com }

\author{Da-Wei Jaw}
\affiliation{%
  \institution{National Taipei University of Technology}
  \department{EE}
}
\email{jdw.daivdjaw@gmail.com}

\author{Shih-Chia Huang}
\affiliation{%
  \institution{National Taipei University of Technology}
  \department{EE}
}
\email{schuang@ntut.edu.tw}

\author{Jenq-Neng Hwang}
\affiliation{%
  \institution{University of Washington}
  \department{EE}
}
\email{hwang@uw.edu}


\begin{abstract}
Existing learning-based atmospheric particle-removal approaches such as those used for rainy and hazy images are designed with strong assumptions regarding spatial frequency, trajectory, and translucency.
However, the removal of snow particles is more complicated because it possess the additional attributes of particle size and shape, and these attributes may vary within a single image.
Currently, hand-crafted features are still the mainstream for snow removal, making significant generalization difficult to achieve.
In response, we have designed a multistage network codenamed \textit{DesnowNet} to in turn deal with the removal of translucent and opaque snow particles.
We also differentiate snow into attributes of translucency and chromatic aberration for accurate estimation. 
Moreover, our approach individually estimates residual complements of the snow-free images to recover details obscured by opaque snow.
Additionally, a multi-scale design is utilized throughout the entire network to model the diversity of snow. As demonstrated in experimental results, our approach outperforms state-of-the-art learning-based atmospheric phenomena removal methods and one semantic segmentation baseline on the proposed Snow100K dataset in both qualitative and quantitative comparisons.
The results indicate our network would benefit applications involving computer vision and graphics.
\end{abstract}

%
%


%
%



\thanks{
Y.-F. Liu is with Viscovery, Taiwan. E-mail: yunfuliu@gmail.com. \\
D.-W. Jaw and S.-C. Huang is with Department of EE, National Taipei University of Technology, Taiwan. E-mail: jdw.davidjaw@gmail.com,~schuang@ntut.edu.tw. \\
J.-N. H is with Department of EE, University of Washington, USA. E-mail: hwang@uw.edu.
}

\begin{teaserfigure}
	\includegraphics[width=\textwidth]{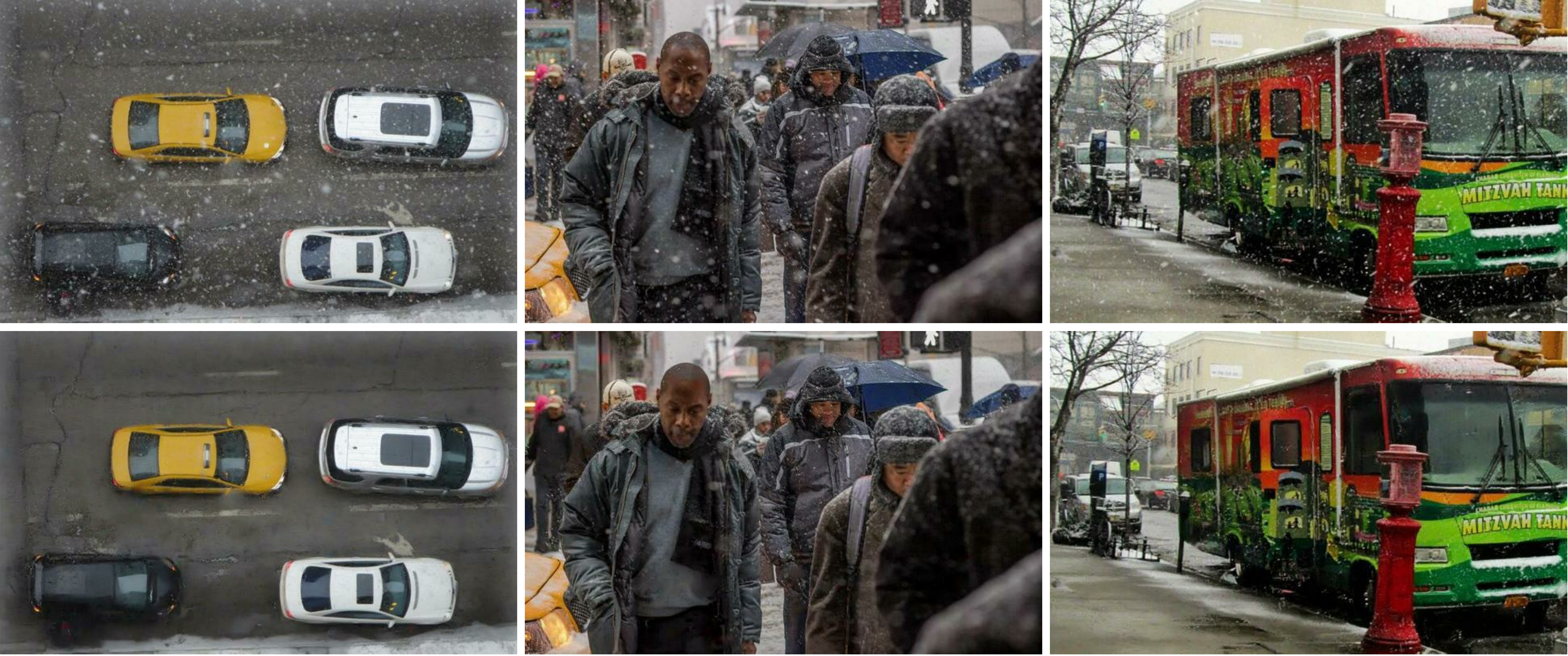}
    \caption{Our \textit{DesnowNet} automatically locates the translucent and opaque snow particles from realistic winter photographs (top row) and removes them to achieve better visual clarity on corresponding resultants (bottom row). Image credits (left to right): Flickr users yooperann, Michael Semensohn, and Robert S.}
\end{teaserfigure}

\maketitle
\renewcommand\thesection{\arabic{section}}
\begin{CJK}{UTF8}{bsmi}

\section{Introduction}

\begin{figure}[t]
  	\centering
    \includegraphics[width=1\linewidth]{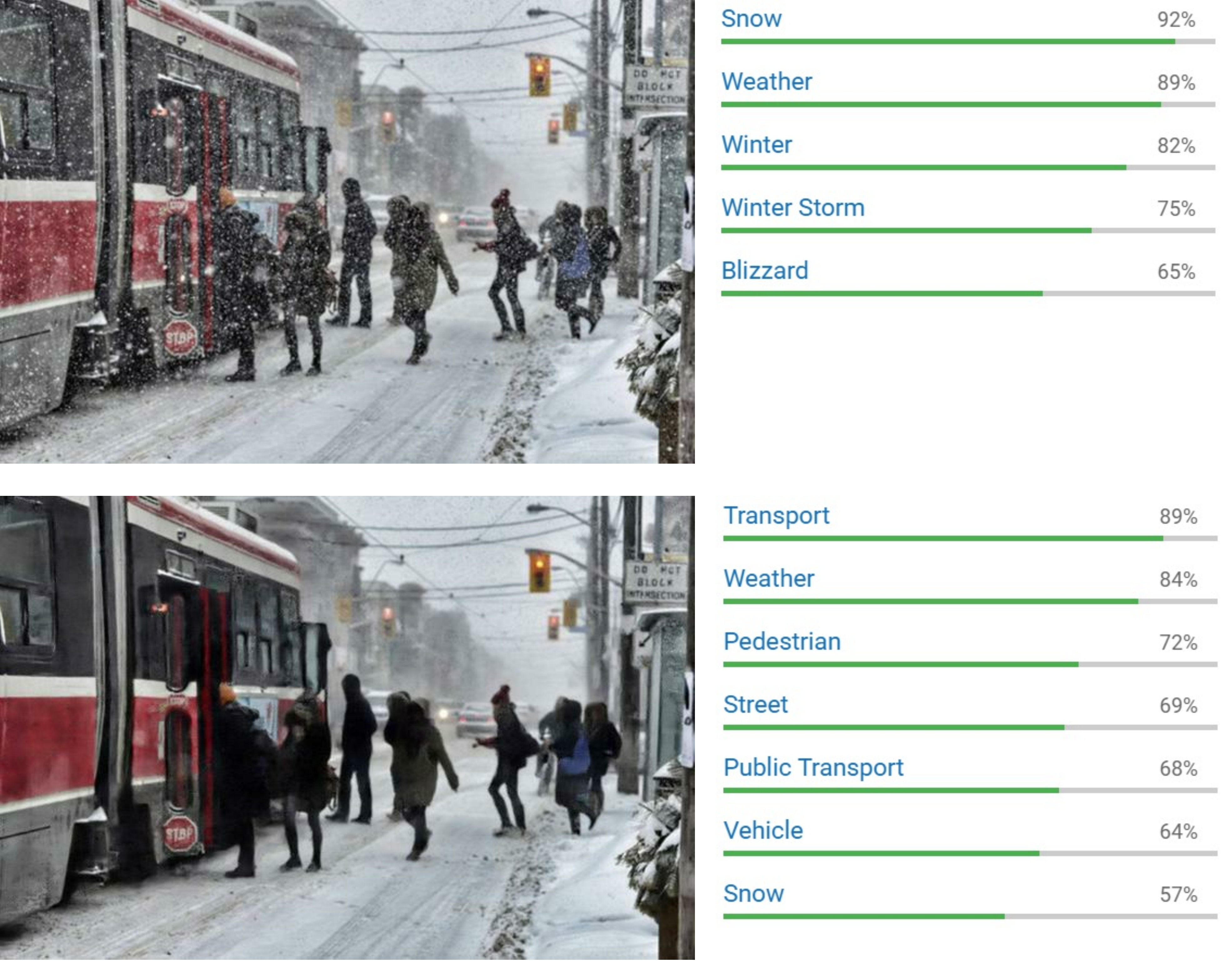}
  	\caption{Realistic winter photograph (top) and corresponding snow removal result (bottom) of the proposed method with labels and confidences supported by Google Vision API.} 
    \label{fig:visionapi}
\end{figure}

Atmospheric phenomena, e.g., rainstorms, snowfall, haze, or drizzle, obstructs the recognition of computer vision applications. Such conditions can influence the sensitive usages such as intelligent surveillance systems and result in higher risks of false alarms and unstable machine interpretation.
Figure \ref{fig:visionapi} shows a subjective but quantifiable result for a concrete demonstration, in which the labels and corresponding confidences are both supported by \href{https://cloud.google.com/vision/}{Google Vision API}\footnote{Google Vision API: https://cloud.google.com/vision/}. 
It illustrates that these atmospheric particles could impede the interpretation of object-centric labeling - in this case, of \textit{pedestrian} and \textit{vehicle}.

Numerous atmospheric particle removal techniques have been proposed to eliminate object obscuration by particles and thus provide richer details.
To this end, removing haze and rain particles from images are common topics of research currently.
So far, approaches for dealing with haze particles have been designed with a strong assumption \cite{tan2008visibility,fattal2008single}, attenuation prior \cite{chen2016edge,he2011single,chen2016high,huang2014efficient,chen2015advanced,huang2015advanced,huang2014visibility} and feature learning \cite{dehaze2016}, based on the observation that haze is uniformly accumulated over an entire image.
Moreover, rain removal techniques aim to model the general characteristics such as edge orientation \cite{kang2012automatic,luo2015removing,chen2014visual}, shapes \cite{derain2016} or patterns \cite{li2016rain} to detect and remove rain particles.
In recent years, learning-based approaches such as DerainNet \cite{derain2017} and DehazeNet \cite{derain2016} have attracted much attention as they are purportedly more effective than the previous hand-crafted features due to significantly improved generalization. 

Although the learning-based rain and haze removal methods are capable of locating and removing atmospheric particles, it is hard to adapt them to snow removal because of the complicated characteristics.
Specifically, its uneven density, diversified particle sizes and shapes, irregular trajectory and transparency make snow removal a more difficult task to accomplish and inapplicable for other learning-based methods. 
Meanwhile, existing snow removal approaches focus on designing hand-crafted features, which often results in weak generalization ability as that which has happened in the fields of rain and haze removal.

In this paper, we design a network named \textit{DesnowNet} to in turn deal with complicated translucent and opaque snow particles. 
The reason behind this multistage design is to acquire more recovered image content by which to accurately estimate and restore details lost to opaque snow particles coverage. 
In addition, we model snow particles with the attributes of a snow mask, which considers only the translucency of snow at each coordinate, as well as an independent chromatic aberration map by which to depict subtle color distortions at each coordinate for better image restoration.
Because of the variations in size and shape of snow particles, the proposed approach comprehensively interprets snow through context-aware features and loss functions.
Experimental results demonstrate that proposed approach yield a significant improvement of prediction accuracy as evaluated via the proposed Snow100K dataset.  

The rest of this paper is organized as follows: Section \ref{sec:relatedworks} provides an overview of the existing atmospheric particle removal methods. Sections \ref{proposed method} and \ref{sec:dataset} elaborate the details of the proposed DesnowNet and Snow100K dataset, respectively. Finally, Section \ref{experiment} presents the experimental results, and Section \ref{conclusion} draws the conclusions.

\section{Related Works} \label{sec:relatedworks}

\subsection{Rain Removal} \label{sec:rainremoval}
Kang et al. \cite{kang2012automatic} proposed the first image decomposition framework for single image rain streak removal.
Their method is based on the assumption that rain streaks have similar gradient orientations in one image, as well as high spatial frequency. These features are considered to segment the rainy component with sparse representation for the removal process.
Luo et al. \cite{luo2015removing} proposed a similar discriminative sparse coding methodology with the same assumption to remove rain particles from rainy images.
However, both methods introduce stripe-like artifacts. 

To cope with this, Son $et~al.$ \cite{derain2016} designed a shrinkage-based sparse coding technique to improve visual quality.
In addition, $Chen~et~al.$ \cite{chen2014visual} extended \cite{kang2012automatic} via the histogram of oriented gradients feature (HOG \cite{dalal2005histograms}), depth of field, and Eigen color to better separate rain components from others.
Li et al. \cite{li2016rain} proposed an image decomposition framework using the Gaussian mixture model to accomplish rain removal by accommodating the similar patterns and orientations among rain particles.

To further improve generalization, Fu et al. \cite{derain2017} most recently proposed a convolutional neural network (CNN)-based method termed DerainNet to remove rain particles.
In their design, they first separate rainy image into a high frequency detail layer and a low frequency base layer, under the assumption that they possess and are devoid of rain particles, respectively. 
This method outperforms the former hand-crafted methods when it comes to removal accuracy and the clarity of the results. 
Although individual cases of snowy images may be similar in appearance to those featuring rain streaks, the most common scenario in which a snowy image features coarse-grained snow particles will result in failure for DerainNet \cite{derain2017} due to its overly focused spatial frequency.

\subsection{Haze Removal} \label{sec:hazeremoval}

Tan's method \cite{tan2008visibility} assumes that haze-free images possess better contrast than images with haze. Thus they remove haze particles by maximizing the local contrast of hazy images.
Fattal $et~al.$ \cite{fattal2008single} removed haze by estimating the albedo of a scene and inferring the transmission medium in hazy image.
Chen $et~al.$ \cite{chen2016edge} proposed a self-adjusting transmission map estimation technique by taking advantage of edge collapse.
He $et~al.$ \cite{he2011single} accomplish haze removal via dark channel prior, attracting considerable attention from their effective results.

Chen $et~al.$ \cite{chen2016high} proposed a gain intervention refinement filter to speed up the runtime of \cite{he2011single}.
In \cite{huang2014efficient}, a hybrid dark channel prior is proposed to avoid the artifacts introduced by localized light sources.
As an extension, the dual dark channel prior with adaptable localized light detection \cite{chen2015advanced} was introduced to automatically adjust the size of a binary mask to conceal localized light sources and achieve more reliable results for haze removal.
Huang $et~al$. \cite{huang2015advanced} proposed a Laplacian-based visibility restoration technique to refine the transmission map and solve color cast issues.
In \cite{huang2014visibility}, a transmission map refinement procedure is proposed to avoid complex structure halo effects by preserving the edge information of the image.

However, all these hand-crafted methods often suffer from similar failure cases due to haze-relevant priors or heuristic cues.
Most recently, Cai $et~al.$ \cite{dehaze2016} proposed a learning-based method termed DehazeNet to learn the mapping between a hazy image and its corresponding medium transmission map. 
In their atmospheric scattering model, they assume that a clear image is an superposition of an equal-brightness haze mask with different translucency and a clear image.
While their method is effective in some cases, there are limitations.
For instance, the brightness of the haze mask is directly determined by the global maximum of the image, which is not learnable and may also cause problems for generalization.
Moreover, their approach also assumes that haze particles are translucent and that the images lack opaque corruption. Such architecture focusing on extracting global features is designed to recover entire image corruption, which leads to the inapplicable on snow removal.

\subsection{Snow Removal} \label{sec:snowremoval}
Unlike the characteristics of atmospheric particles in rainy and hazy images that might be described as relatively similar in spatial frequency, trajectory, and translucency, the variations in the particle shape and size of snow makes it more complex. 
However, existing snow removal methods inherited priors of rainfall-driven features, e.g., HOG \cite{bossu2011rain,pei2014removing}, frequency space separation \cite{rajderkar2013removing} and color assumptions \cite{xu2012improved,xu2012removing} to model falling snow particles.
These features not only model just the partial characteristics of snow, but worsen the prospects of generalization.


\section{Proposed Method} \label{proposed method}

We now describe our proposed method for removing falling snow particles from snowy images. 
Suppose that a snowy color image $\mathbf{x} \in [0, 1]^{p \times q \times 3}$ of size $p \times q$ is composed of a snow-free image $\mathbf{y} \in [0, 1]^{p \times q \times 3}$ and an independent snow mask $\mathbf{z} \in [0, 1]^{p \times q \times 1}$ which introduces only the translucency of snow.
This relationship can be described as:
\begin{equation} \label{eq-overlap}
\mathbf{x} = \mathbf{a} \odot \mathbf{z} + \mathbf{y} \odot (1-\mathbf{z})
\end{equation}
where $\odot$ denotes element-wise multiplication, and the chromatic aberration map $\mathbf{a} \in \mathbb{R}^{p \times q \times 3}$ introduces the color aberration at each coordinate.

\begin{figure}[t]
	\centering
    \includegraphics[width=0.3\textwidth]{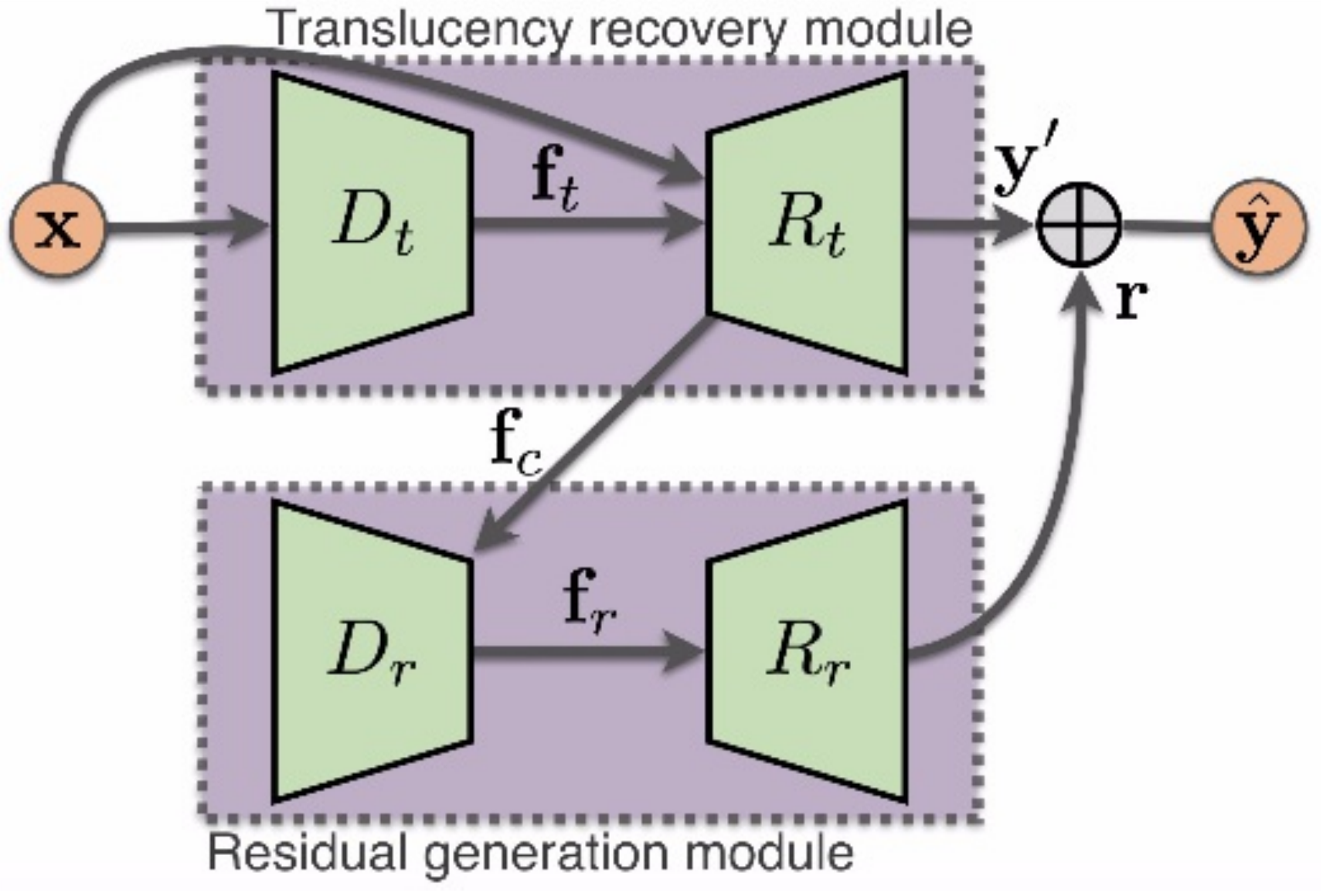}
  	\caption{Overview of the proposed DesnowNet.}
    \label{fig:DSN}
\end{figure}

\begin{figure}[t]
	\begin{center}
      \begin{subfigure}{.23\textwidth}
        \centering
        \includegraphics[width=.97\linewidth]{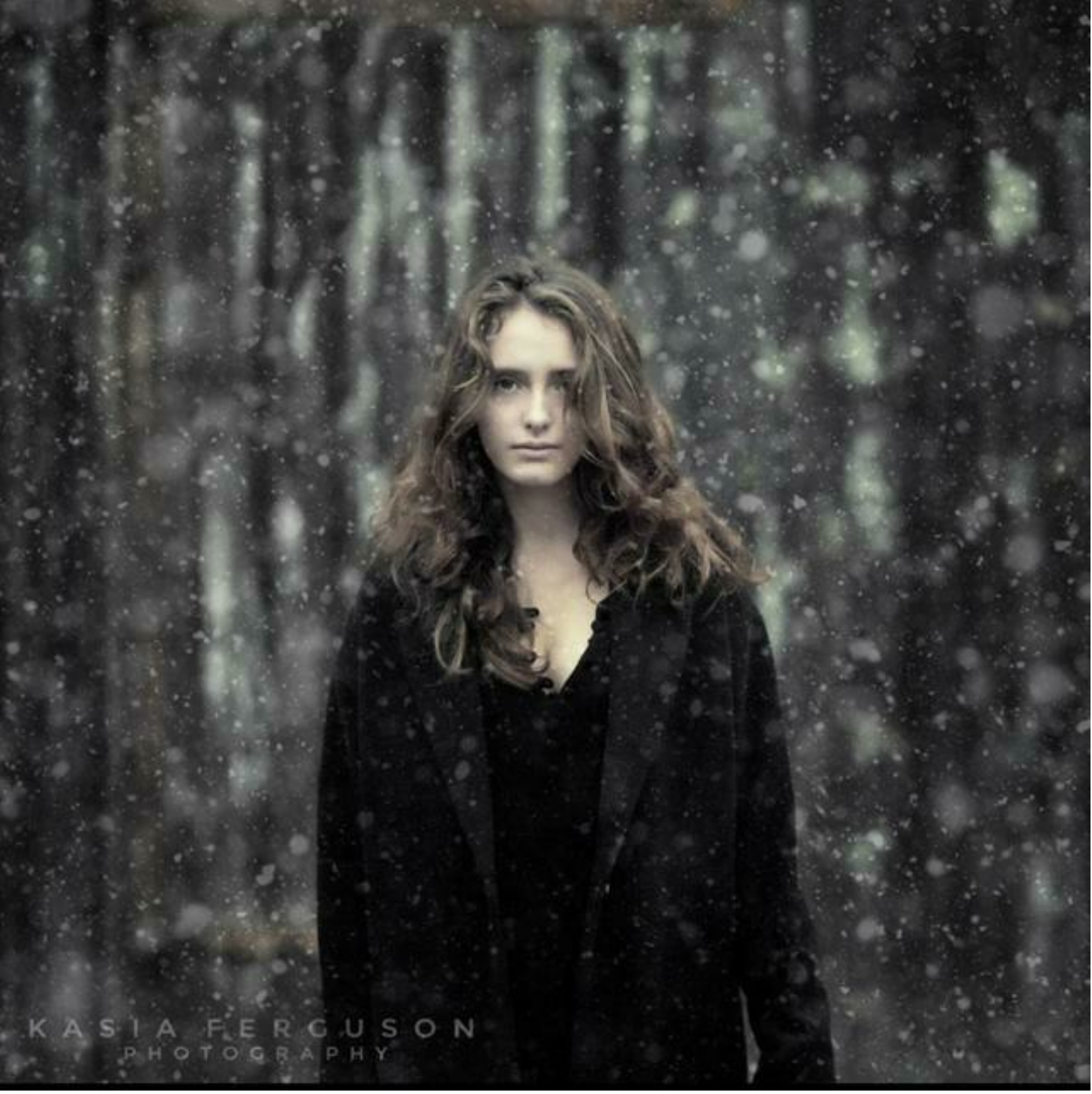}
        \caption{Snowy image ($\mathbf{x}$)}
      \end{subfigure}%
	\end{center}    
	\begin{center}
      \begin{subfigure}{.23\textwidth}
        \centering
        \includegraphics[width=0.97\linewidth]{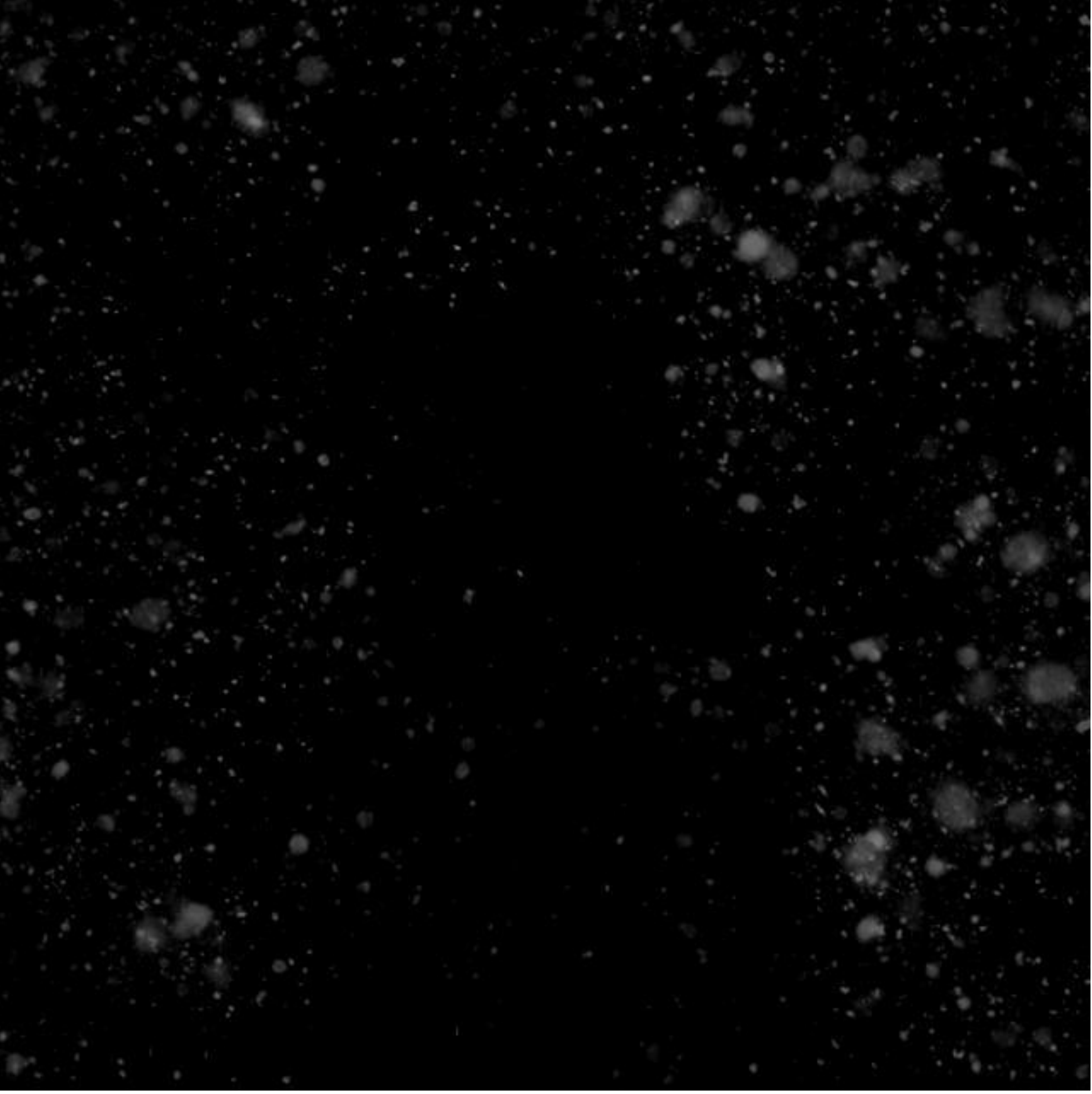}
        \caption{Aberrated snow mask ($\mathbf{a} \odot \mathbf{\hat{z}}$)}
      \end{subfigure}%
      \begin{subfigure}{.23\textwidth}
        \centering
        \includegraphics[width=0.97\linewidth]{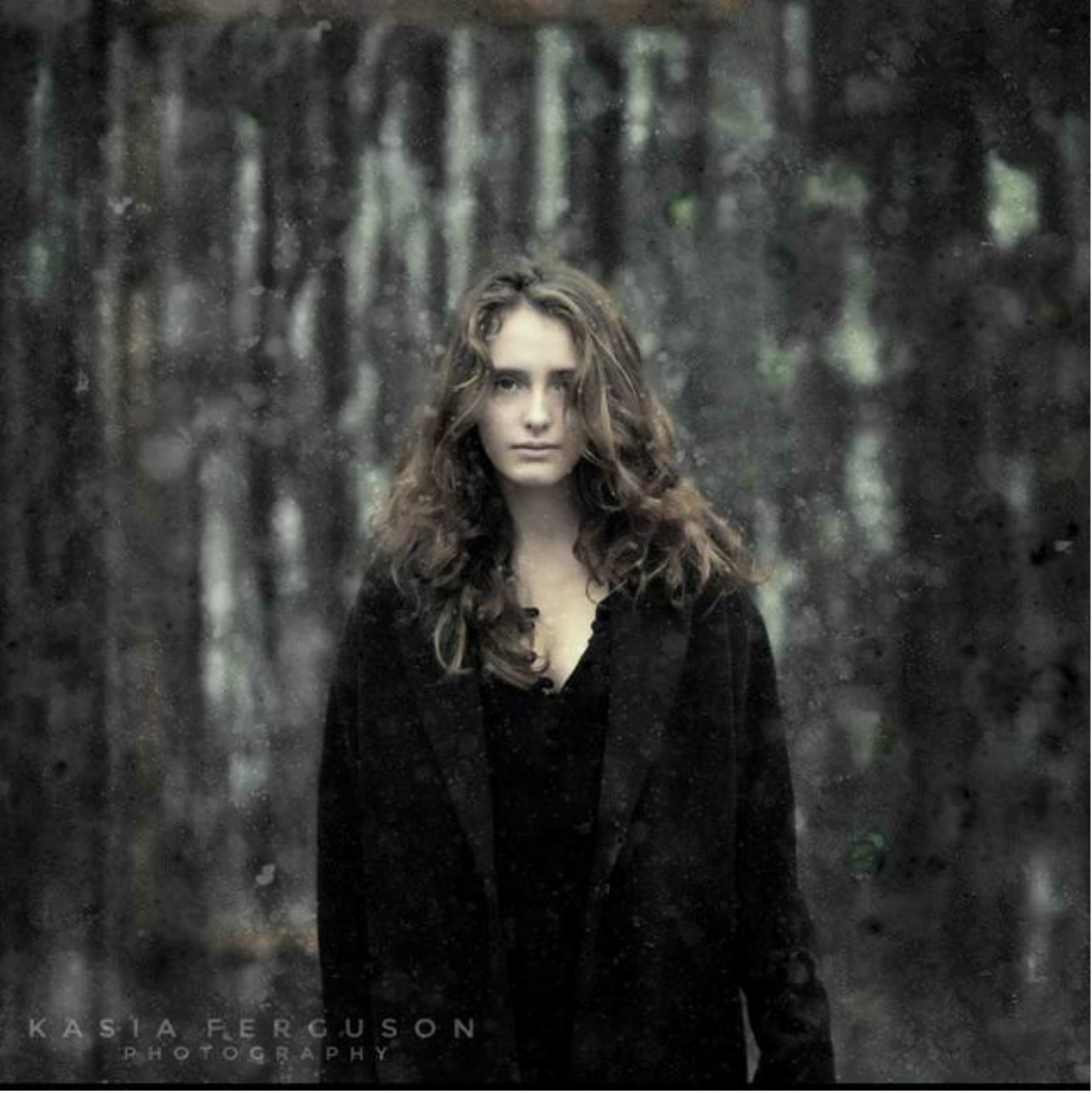}
        \caption{Estimated snow-free output ($\mathbf{y}'$)}
      \end{subfigure}%
	\end{center}
	\begin{center}
      \begin{subfigure}{.23\textwidth}
        \centering
        \includegraphics[width=0.97\linewidth]{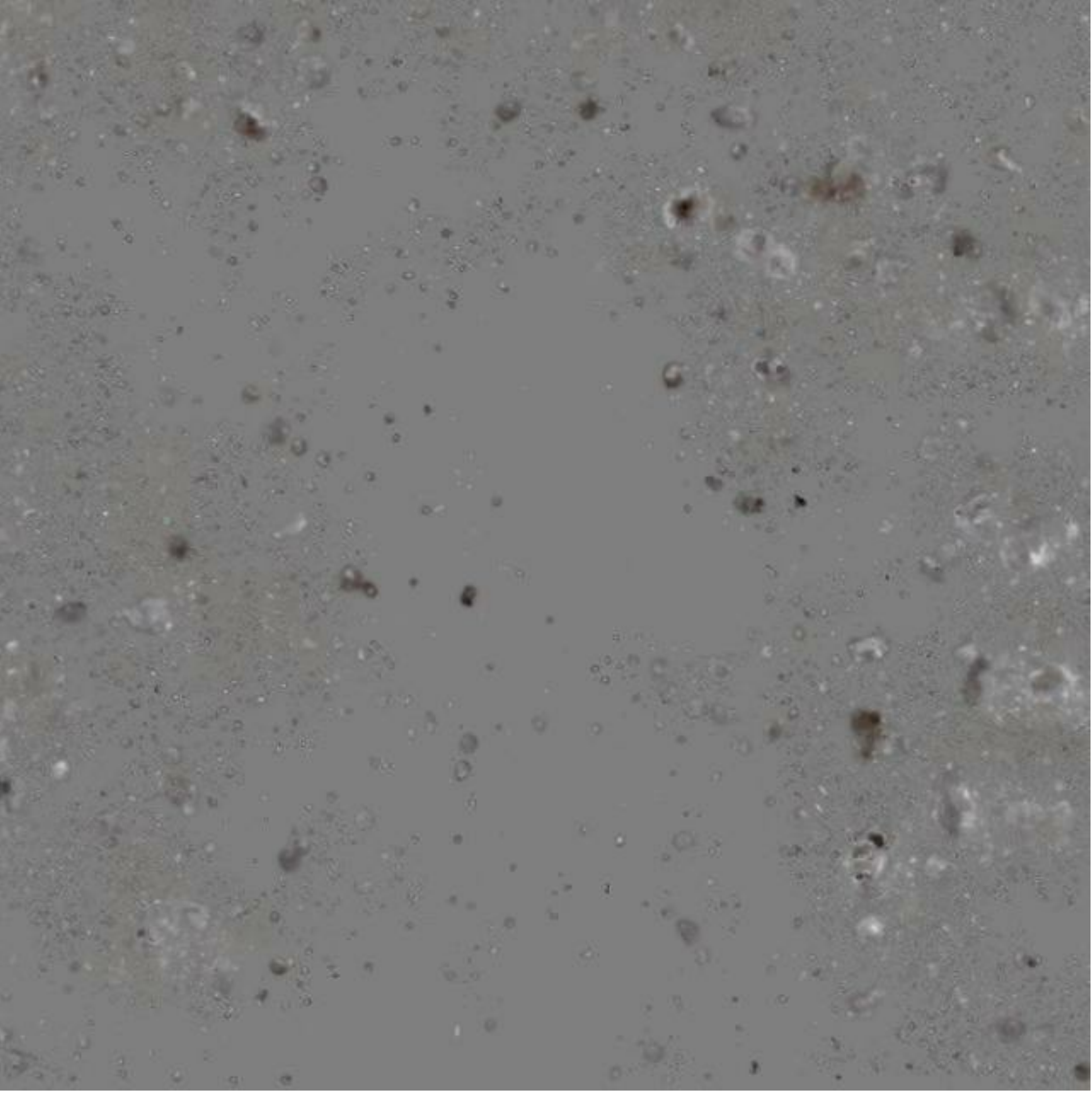}
        \caption{Residual complement ($\mathbf{r}$)}
      \end{subfigure}%
      \begin{subfigure}{.23\textwidth}
        \centering
        \includegraphics[width=0.97\linewidth]{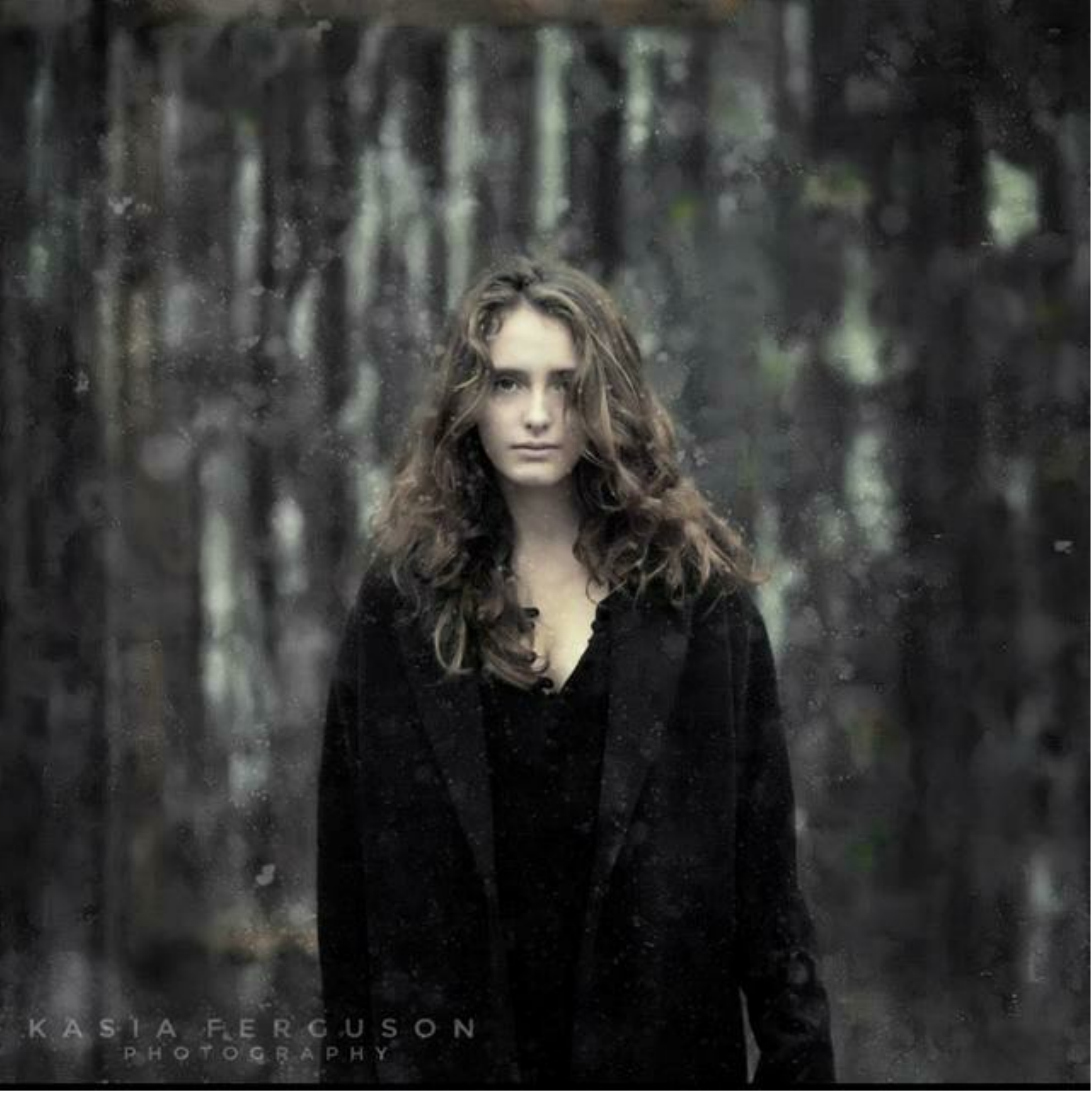}
        \caption{Estimated snow-free output ($\mathbf{\hat{y}}$)}
      \end{subfigure}%
  	\caption{Results of DesnowNet during inference, where the gray color in (d) denotes $r = 0$, and white and black colors represent positive and negative values, respectively, where $r \in \mathbf{r}$.}        
	\label{fig:results_flow}
	\end{center}    
\end{figure}

To derive the estimated snow-free image $\hat{\mathbf{y}}$ from a given $\mathbf{x}$, the estimation of an accurate snow mask $\mathbf{\hat{z}}$ as well as a chromatic aberration map $\mathbf{a}$ are crucial to achieve an appealing visual quality.
To this end, we design a network with multi-scale receptive fields to model all of the variations of snow particles as introduced in Section \ref{sec:snowremoval}. 
Specifically, the proposed network as depicted in Fig. \ref{fig:DSN} consists of two modules for different purposes: 
1) the translucency recovery (TR) module, which recovers areas obscured by translucent snow  particles; 
2) the residual generation (RG) module, which generates the residual complement $\mathbf{r} \in \mathbb{R}^{p \times q \times 3}$ of the estimated snow-free image $\mathbf{y}'$ for portions completely obscured by opaque snow particles according to the clear (non-covered) areas as well as those recovered by the TR module. 
To do so, two modulized descriptors $D_t$ and $D_r$ are utilized to extract multi-scale features $\mathbf{f}_t$ and $\mathbf{f}_r$ 
with varied receptive fields, and two modulized recoverers $R_t$ and $R_r$ are designed to yield the snow-free estimate $\mathbf{y}'$ and residual complement $\mathbf{r}$. Thus, the estimated snow-free result ($\hat{\mathbf{y}}$) can be derived as described below:
\begin{equation} \label{eq:y_hat}
\mathbf{\hat{y}} = \mathbf{y}' + \mathbf{r}
\end{equation}
The derivation of $\mathbf{y}'$ and $\mathbf{r}$ are further described in the following subsections. 
Notably, due to the possibility that the value range of $\mathbf{r}$ could make $\mathbf{\hat{y}}$ exceed the expected range $[0, 1]$, we clip when $\hat{y}_i > 1$ and $\hat{y}_i < 0$ during inference yet keep it unchanged during training, where $\hat{y}_i \in \mathbf{\hat{y}}$.   
Fig. \ref{fig:results_flow} illustrates examples of the variables in Fig. \ref{fig:DSN} to aid comprehension of this process.

\subsection{Descriptor} \label{sec:descriptor}

Both $D_t$ and $D_r$ (illustrated in Fig. \ref{fig:DSN}) utilize the same type of descriptor to introduce the variations of snow particles. Specifically, although the purposes of $D_t$ and $D_r$ are inherently different as described above, they share the same need for resolving the variations in snow particles. 
To this end, we first employ Inception-v4 \cite{DBLP:inceptionv4} as the backbone due to its highly optimized features at multi-scale receptive fields, a network whose success has been demonstrated in diverse studies \cite{brock2016neural-inception,diamond2017dirty-inception}

To further extend context-awareness and exploit multi-scale features, we proposed a new subnetwork termed dilation pyramid (DP) inspired by the atrous spatial pyramid pooling (ASPP) in DeepLab \cite{deeplab}. It further enhances the capability of extracting features at scaling invariant, which is also the apparent variation in falling snows. 

Instead of directly summing up the collected multi-scale features, as in ASPP, which might lose important spatial information, we lean to preserve it by concatenating the features as formulated below:
\begin{equation} \label{eq:DP}
\mathbf{f}_t = \bigparallel^\gamma_{n=0}B_{2^n}(\Phi(\mathbf{x}))
\end{equation}
where $\Phi(\mathbf{x})$ represents the network of Inception-v4 with a given image $\mathbf{x}$, $B_{2^n}(\cdot)$ denotes the dilated convolution \cite{yu2015multi} (the atrous convolution in \cite{deeplab})  with dilation factor $2^n$, $\bigparallel$ represents the concatenating operation, $\gamma \in \mathbb{R}$ denotes the levels of the dilation pyramid, and $\mathbf{f}_t$ (or $\mathbf{f}_r$ for $D_r$ in Fig. \ref{fig:DSN}) denotes the output feature of $D_t$. 

\subsection {Recovery Submodule} \label{sec:recovery-module}

\begin{figure}[t]
	\centering
    \includegraphics[width=0.25\textwidth]{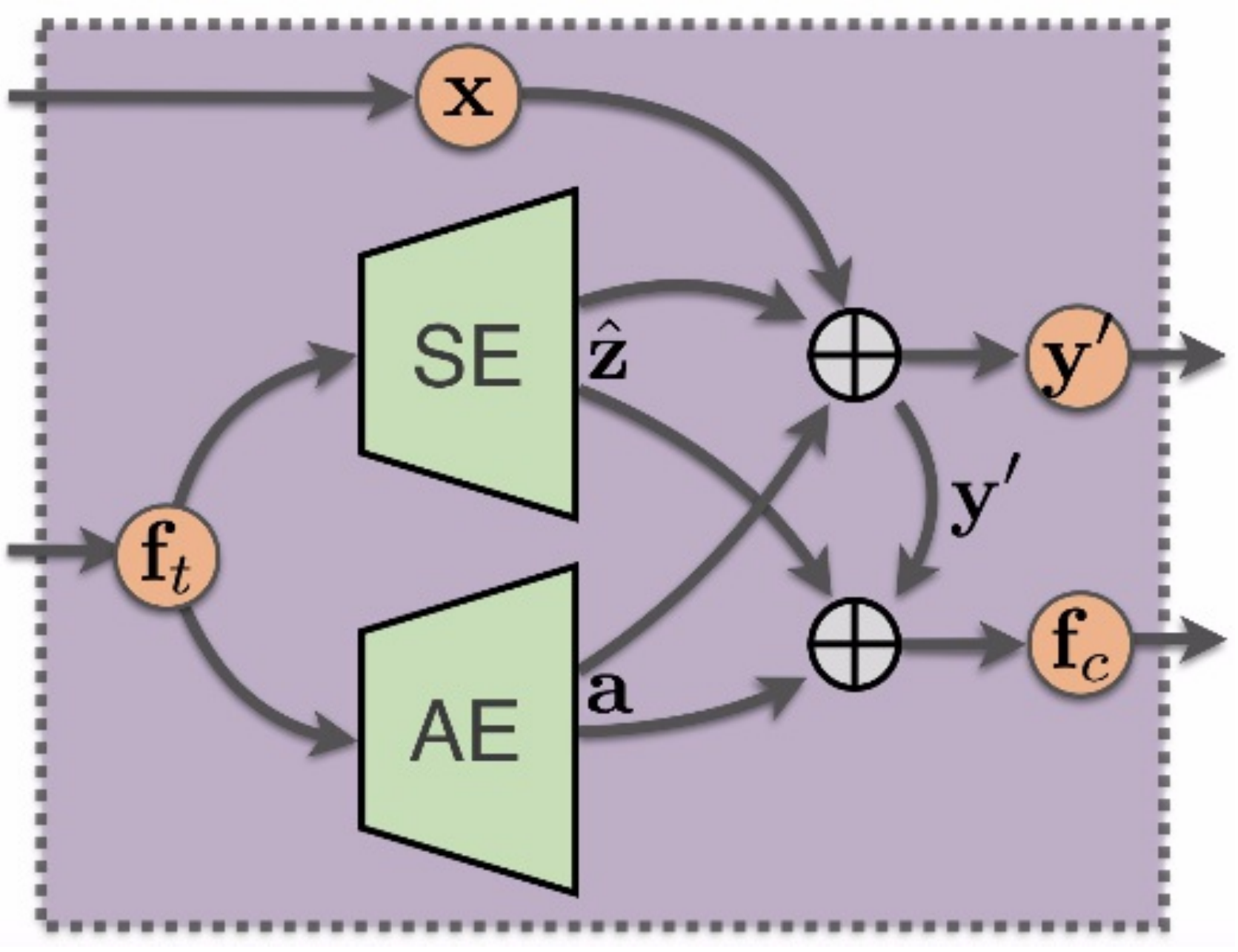}
  	\caption{Overview of the $R_t$ submodule.}
    \label{fig:R_t}
\end{figure}

\textbf{Pyramid maxout.}
The recovery submodule ($R_t$) of the translucency recovery (TR) module illustrated in Fig. \ref{fig:DSN} generates the estimated snow-free output ($\mathbf{y}'$) by recovering the details behind translucent snow particles. 
To accurately model this variable, we physically and separately define it as two individual attributes as depicted in Fig. \ref{fig:R_t}: 
1) snow mask estimation (SE), which generates a snow mask ($\hat{\mathbf{z}} \in [0, 1]^{p \times q \times 1}$) to model the translucency of snow with a single channel, and 
2) aberration estimation (AE), which generates the chromatic aberration map ($\mathbf{a} \in \mathbb{R}^{p \times q \times 3}$) to estimate the color aberration of each RGB channel. 
Their relationship is defined in Eq. (\ref{eq-overlap}). 
Due to the variation in size and shape of snow particles, we design an emulation architecture termed $pyramid~maxout$, which is defined below, to enforce the network to select the robust feature map from different receptive fields as its output.
It is similar to the general form of maxout \cite{goodfellow2013maxout} in terms of its competitive policy, and is defined as:
\begin{equation} \label{eq:pyramid_maxout}
M_\beta(\mathbf{f}_t) = max(\text{conv}_1(\mathbf{f}_t), \text{conv}_3(\mathbf{f}_t), \ldots, \text{conv}_{2\beta-1}(\mathbf{f}_t))
\end{equation}
where $\text{conv}_n(\cdot)$ denotes the convolution with kernel of size $n \times n$, $max(\cdot)$ denotes the element-wise maximum operation, $\mathbf{f}_t$ has been defined in Eq. (\ref{eq:DP}), and parameter $\beta \in \mathbb{R}$ controls the scale of the pyramid operation. 
In our design, SE and AE utilize the same architecture as defined in Eq. (\ref{eq:pyramid_maxout}) to generate $\mathbf{\hat{z}}$ and $\mathbf{a}$, respectively. 

\begin{figure}[t]
	\begin{center}
      \begin{subfigure}{.4\textwidth}
        \centering
        \includegraphics[width=.97\linewidth]{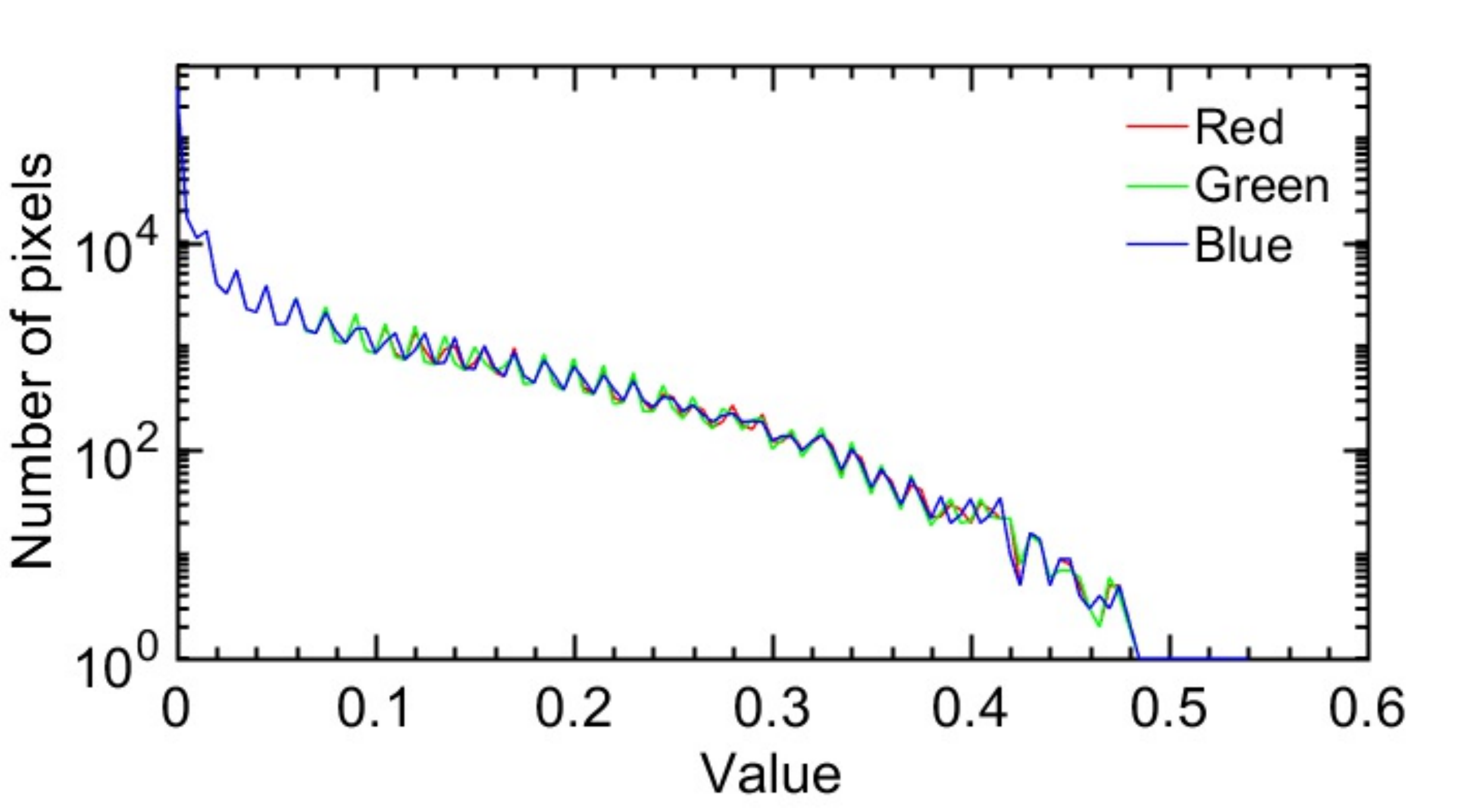}
        \caption{Aberrated snow mask ($\mathbf{a} \odot \mathbf{\hat{z}}$)}
      \end{subfigure}%
	\end{center}
	\begin{center}
      \begin{subfigure}{.4\textwidth}
        \centering
        \includegraphics[width=.97\linewidth]{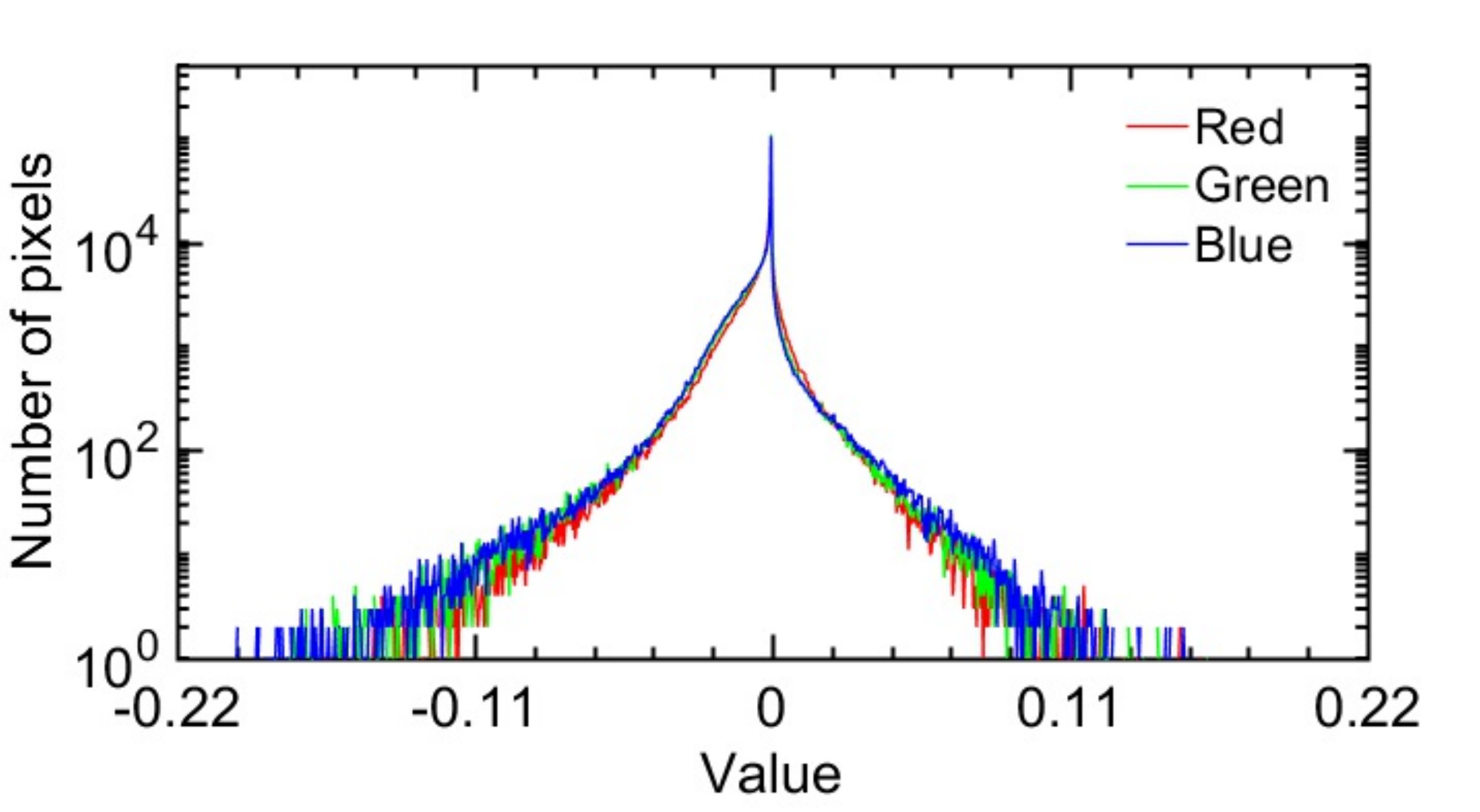}
        \caption{Residual complement ($\mathbf{r}$)}
      \end{subfigure}%
  	\caption{Histograms of the results of Fig. \ref{fig:results_flow}(b) and (d), where RGB represents the color channels.}    
	\label{fig:results_hists}    
	\end{center}    
\end{figure}

\textbf{Translucency recovery (TR).} 
$R_t$ in the TR module recovers the content behind translucent snows to yield an estimated snow-free image $\mathbf{y}'$. The relationship can be formulated with Eq. (\ref{eq-overlap}):
\begin{equation} \label{eq:recovery}
{y}'_i = 
\begin{cases}
\frac{{x}_i - {a}_i \times {\hat{z}}_i}{1 - {\hat{z}}_i}, & \text{if }{\hat{z}}_i < 1 \\
{x}_i, & \text{if } {\hat{z}}_i = 1
\end{cases}
\end{equation}
where the subscript $i$ in $\hat{z}_i \in \hat{\mathbf{z}}$, $a_i \in \mathbf{a}$, $x_i \in \mathbf{x}$, and $y'_i \in \mathbf{y}'$ denote the $i$-th element in corresponding matrices. Notably, a condition of $y_i' = x_i$ occurs if $\hat{z}_i = 1$ addresses the spatial case of nonexistent $y_i'$ as defined in Eq. (\ref{eq-overlap}). 
Fig. \ref{fig:results_flow}(b) and \ref{fig:results_hists}(a) show the results of $\mathbf{a} \odot \mathbf{\hat{z}}$ as defined in Eq. (\ref{eq:recovery}). 
Although the snow particles are normally presented in middle gray in most photographs, such as that shown in Fig. \ref{fig:results_flow}(b), our chromatic aberration map $\mathbf{a}$ describes  potential subtle color variations as depicted in Fig. \ref{fig:results_hists}(a). This not only further complements the property of colored snow particles, but improves generalization of variation in ambient lights.

\textbf{Residual generation (RG).}
$R_r$ serves a different purpose that generates a residual complement $\mathbf{r} \in \mathbb{R}^{p \times q \times 3}$ for the estimated snow-free image $\mathbf{y}' \in \mathbb{R}^{p \times q \times 3}$ to yield a visually appealing $\hat{\mathbf{y}}$ as formated in Eq. (\ref{eq:y_hat}). 
To this end, the RG module needs to know the explicit locations of estimated snow particles in $\hat{\mathbf{z}}$, corresponding chromatic aberration map $\mathbf{a}$, and the recovered output $\mathbf{y}'$.
The residual complement can be formulated as:
\begin{equation} \label{eq:r}
\begin{aligned}
\mathbf{r} & = R_r(D_r(\mathbf{f}_c)) \\
& = \Sigma_\beta(\mathbf{f_r}) = \sum_{n=1}^\beta \text{conv}_{2n - 1}(\mathbf{f_r}) \end{aligned}
\end{equation}
where $\mathbf{f}_c = \mathbf{y}' \bigparallel \mathbf{\hat{z}} \bigparallel \mathbf{a}$, $\mathbf{f}_r$ denotes the output of $D_r$ as depicted in Fig. \ref{fig:DSN}, $\beta$ is identical to that of in Eq. (\ref{eq:pyramid_maxout}), and pyramid sum $\Sigma_\beta(\cdot)$ aggregates the multi-scale features to model the variation in snow particles. 

It is worth noting that, we do not consider the pyramid maxout as defined in Eq. (\ref{eq:pyramid_maxout}) useful for predicting $\mathbf{r}$ for two reasons: 
1) the distribution of $\mathbf{r}$ should be a zero-mean distribution as shown in Fig. \ref{fig:results_hists}(b) to fairly compensate for the estimated snow-free result $\mathbf{y}'$, and the pyramid maxout would violate this property; 
2) pyramid maxout focuses on the robustness of the estimated features over various scales, whereas the pyramid sum aims to simulate visual perception at all scales simultaneously.

\subsection{Loss Function}

Although the primary purpose of this work aims to remove falling snows from snowy images ($\mathbf{x}$) and thereby approach the snow-free ground truth ($\mathbf{y}$), perceptual loss is needed to simulate its visual similarity. 
Element-wise Euclidean loss is often considered for this purpose. However, it simply evaluates feature representation at single scale constraints for the simulation of various viewing distances pertinent to human vision, and unfortunately introduces unnatural artifacts.
To address this issue, Johnson et al. \cite{johnson2016perceptual} constructed a loss network that measured losses at certain layers. 
We adopt this same idea, retaining its contextual features while formulating it as a lightweight pyramid loss function as defined below:
\begin{equation} \label{eq:pyramid_loss}
\mathfrak{L}(\mathbf{m}, \mathbf{\hat{m}}) = \sum^{\tau}_{i=0} \Vert P_{2^i}(\mathbf{m}) - P_{2^i}(\mathbf{\hat{m}}) \Vert_2^2
\end{equation}
where $\mathbf{m}$ and $\mathbf{\hat{m}}$ denote two images of the same size, $\tau \in \mathbb{R}$ denotes the level of loss pyramid, $P_{n}$ denotes the max-pooling operation with kernel size $n \times n$ and stride $n \times n$ for non-overlapped pooling.

Two feature maps $\mathbf{y}'$ and $\mathbf{\hat{y}}$ represent the estimated snow-free images as illustrated in Fig. \ref{fig:DSN}. In addition, the individual estimated snow mask $\mathbf{\hat{z}}$ depicted in Fig. \ref{fig:R_t} represents the perceptual translucency of snows in the ideal snow-free image $\mathbf{y}$. Hence, the overall loss function is defined as: 
\begin{equation} \label{eq:overall_loss}
\mathcal{L}_{overall} = \mathcal{L}_{\mathbf{y}'} +  \mathcal{L}_{  \hat{\mathbf{y}}} +  \lambda_{\mathbf{\hat{z}}} \mathcal{L}_{\hat{\mathbf{z}}}  + \lambda_{\mathbf{w}} \Vert \mathbf{w} \Vert_2^2
\end{equation}
where $\lambda_{\mathbf{\hat{z}}} \in \mathbb{R}$ denotes the weighting to leverage the importance of snow mask $\mathbf{\hat{z}}$ and both snow-free estimates $\mathbf{y}'$ and $\mathbf{\hat{y}}$, where $\mathbf{y}'$ and $\mathbf{\hat{y}}$ are set at equal importance; 
$\lambda_{\mathbf{w}} \in \mathbb{R}$ denotes the weighting to the $l$2-norm regularization, $\mathbf{w}$ denotes the weighting of the entire DesnowNet; the losses of $\mathbf{y}'$, $\mathbf{\hat{y}}$, and $\mathbf{\hat{z}}$ are defined as:
\begin{equation} \label{eq:loss_terms}
\begin{split}
\mathcal{L}_{\mathbf{\hat{z}}} = \mathfrak{L}(\mathbf{z}, \mathbf{\hat{z}}), 
\mathcal{L}_{\mathbf{\hat{y}}} = \mathfrak{L}(\mathbf{y}, \mathbf{\hat{y}}), \text{and } 
\mathcal{L}_{\mathbf{y}'} = \mathfrak{L}(\mathbf{y}, \mathbf{y}'), 
\end{split}
\end{equation}
where $\mathbf{z}$ and $\mathbf{y}$ denote the ground truth of the snow mask and the ideal snow-free image, and $\mathfrak{L}(\cdot)$ was defined in Eq. (\ref{eq:pyramid_loss}). 


\section{Dataset} \label{sec:dataset}
Due to a lack of availability of public datasets for snow removal, we constructed a dataset termed \href{https://goo.gl/BrRc3U}{Snow100K}\footnote{Snow100K dataset: https://goo.gl/BrRc3U} for the purposes of training and evaluation. 
The dataset $\{(\mathbf{x}_i, \mathbf{y}_i, \mathbf{z}_i)\}^N$ consists of 1) 100k synthesized snowy images ($\mathbf{x}_i$), 2) corresponding snow-free ground truth images ($\mathbf{y}_i$) and 3) snow masks ($\mathbf{z}_i$), and 4) 1,329 realistic snowy images.The images of 2) and 4) were downloaded via the Flickr api, and were manually divided into snow and snow-free categories, respectively. 
In addition, we normalized the size of the largest boundary of each image to 640 pixels and retained its original aspect ratio. 

\begin{figure}[t]
	\begin{center}
      \begin{subfigure}{.158\textwidth}
        \centering
        \includegraphics[width=.97\linewidth]{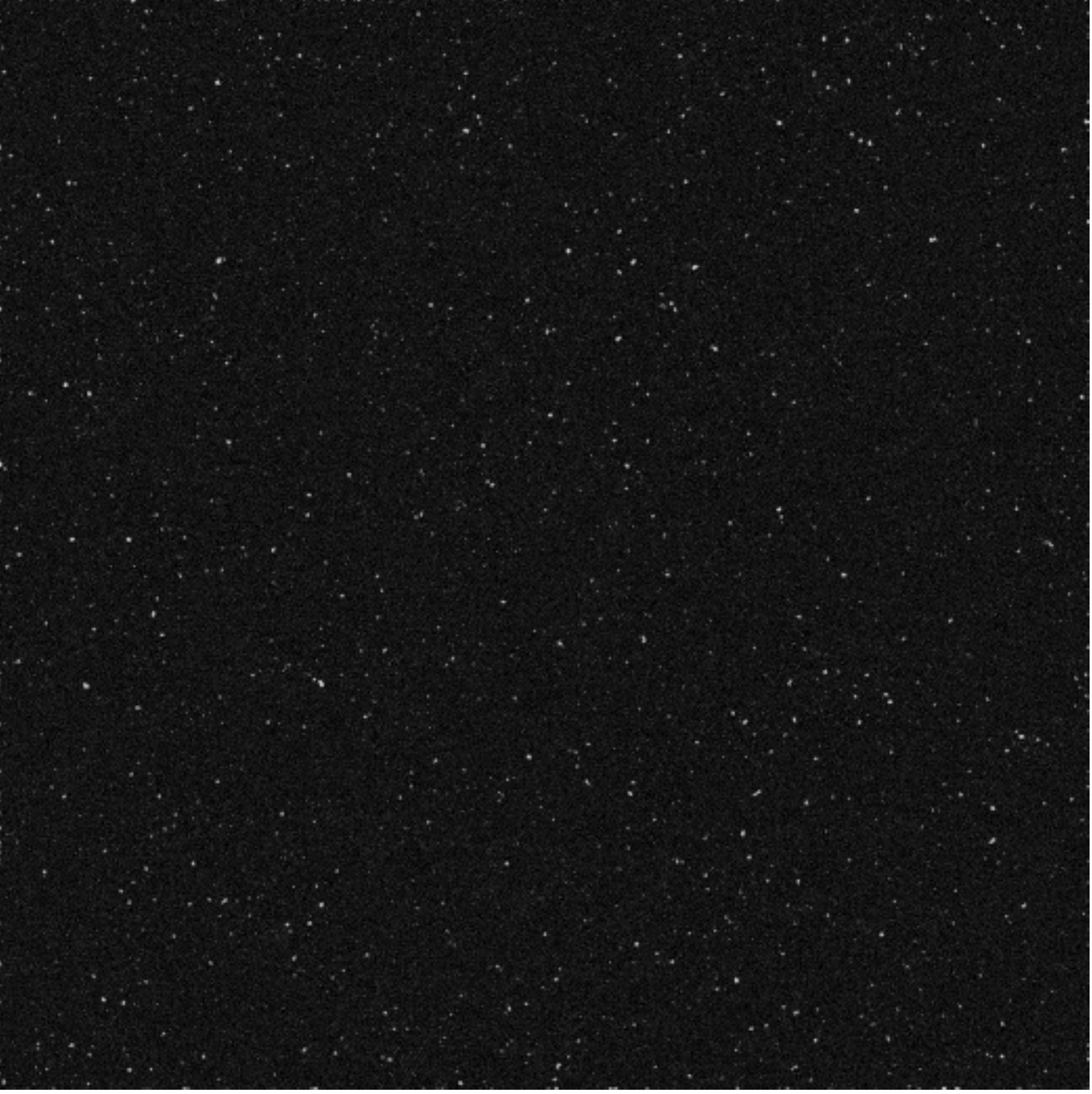}
        \caption{Small}
      \end{subfigure}%
      \begin{subfigure}{.158\textwidth}
        \centering
        \includegraphics[width=.97\linewidth]{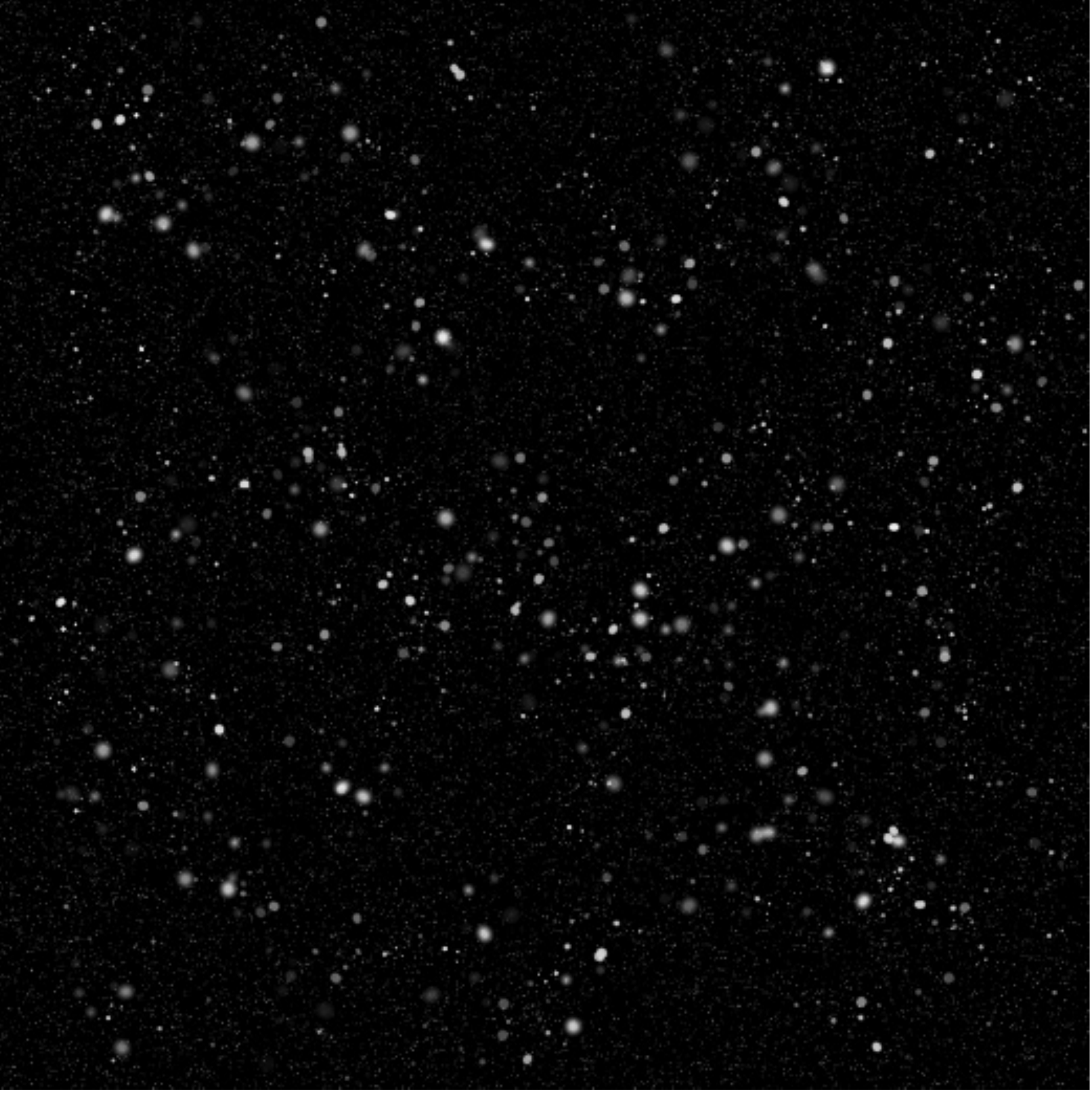}
        \caption{Medium}
      \end{subfigure}%
      \begin{subfigure}{.158\textwidth}
        \centering
        \includegraphics[width=.97\linewidth]{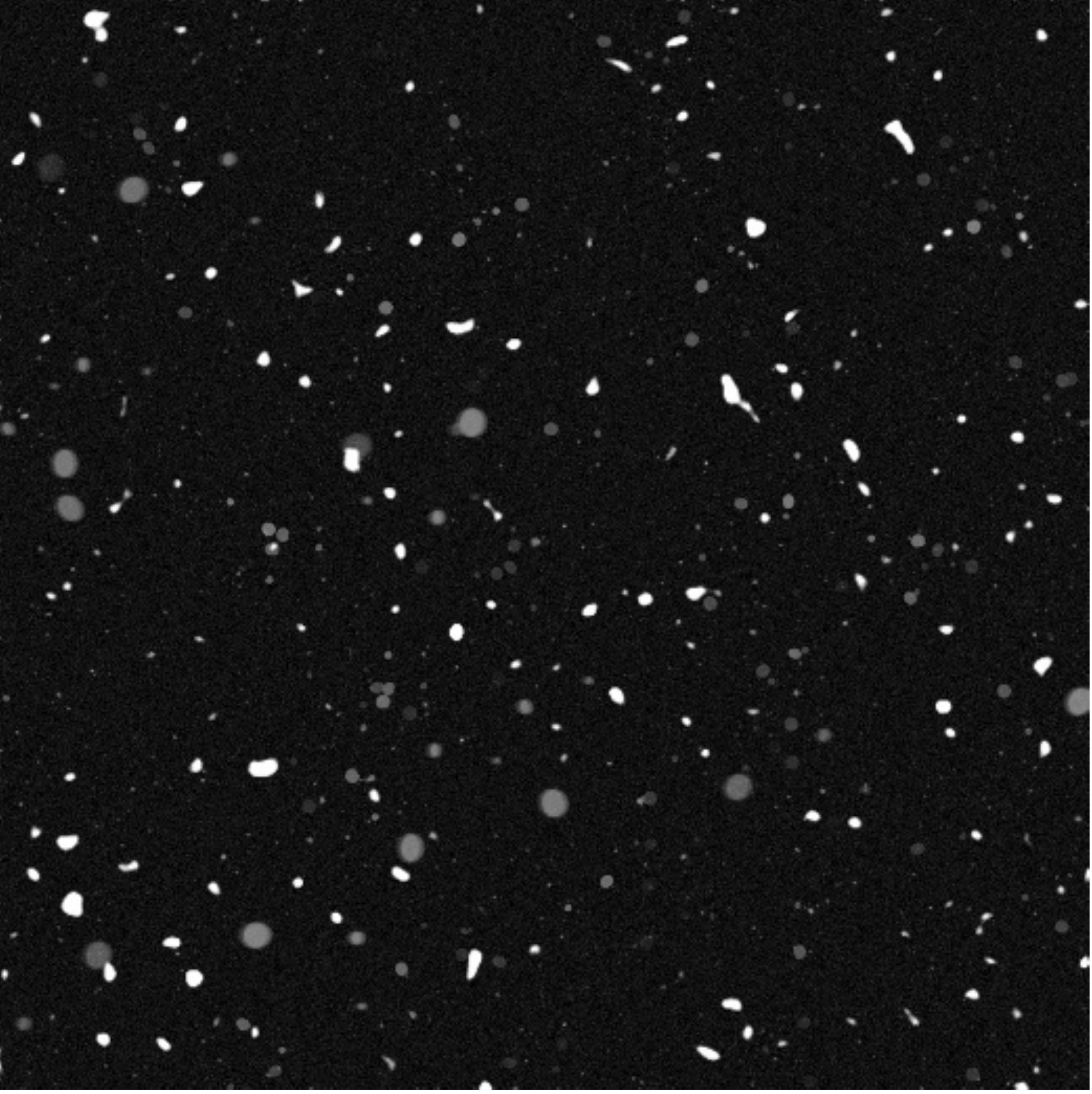}
        \caption{Large}
      \end{subfigure}%
  	\caption{Example of base masks with differing particle sizes.}
    \label{fig:snow_masks}
	\end{center}
\end{figure}

To synthesize the snowy images, we produced 5.8k base masks via PhotoShop to simulate the variation in the particle sizes of falling snow. Each base mask consists of one of the small, medium, and large particle sizes. 
Meanwhile, the snow particles in each base mask feature different densities, shapes, movement trajectories, and transparencies, by which to increase the variation.
The number of base masks in each category is exhibited in Table \ref{table:snow-masks}. 
Corresponding examples are shown in Fig. \ref{fig:snow_masks}.

{\renewcommand{\arraystretch}{1.3}
\begin{table}[t]
\begin{center}
\begin{tabular}{|c|c|c|c|c|}
\hline
Particle size & Small & Medium & Large \\ \hline \hline
\# masks  & 3,800  & 1,600   & 400   \\ \hline
\end{tabular}
\end{center}
\caption{Number of base masks in Snow100K dataset.}
\label{table:snow-masks}
\end{table}
}

Three subsets of synthesized snowy images $\mathbf{x}_i$ are organized for different snowfalls.
Notably, the base masks in Table \ref{table:snow-masks} are utilized here for simulation:
1) Snow100K-S: samples are only overlapped with one randomly selected base mask from the \textit{Small} category\;
2) Snow100K-M: samples are overlapped with one randomly selected base mask from the \textit{Small} category and one base mask from \textit{Medium} category;
3) Snow100K-L: three base masks randomly selected from each of the three categories are adopted to synthesize samples in this subset. 
The numbers of synthesized snowy samples in each category are exhibited in Table \ref{table:subset_count}.
For the superposition process, we append two additional random factors to further increase the randomness for generalization. These include
1) snow brightness: The brightness of the superposed snow in base mask is randomly set within $[max(\mathbf{y}_i) \times 0.7, max(\mathbf{y}_i)]$; and
2) Random cropping: Since the size of base mask is larger than that of the snow-free ground truth $\mathbf{y}_i$, we randomly crop a patch of the base mask with the same size as that of $\mathbf{y}_i$ and superpose the patch over $\mathbf{y}_i$.

{\renewcommand{\arraystretch}{1.3}
\begin{table}[t]
\begin{center}
\begin{tabular}{|c|c|c|c|c|}
\hline
Subset & Snow100K-S & Snow100K-M & Snow100K-L \\ \hline \hline
Training  & 16,643     & 16,622     & 16,735     \\ \hline
Test   & 16,611     & 16,588     & 16,801     \\ \hline
\end{tabular}
\end{center}
\caption{Number of samples in each subset of Snow100K dataset.}
\label{table:subset_count}
\end{table}
}

We then conduct a quantitative experiment to evaluate the reliability of the synthesized snowy images $\mathbf{x}$. As previously mentioned, ground truths are images without snow; yet the conditions in which these images were captured might be unsuitable for our purposes ( e.g., photography of an indoor environment or a shot in outdoor but with sun in the sky) .
To conduct a fair comparison, 500 semantically reasonable synthesized snowy images were collected for the following survey. 
To do so, 30 randomly selected snowy images are shown in each survey and we asked a subject to determine whether each of the given samples is synthetic or realistic one. 
Because of this random selection policy, the numbers of synthesized and realistic images in each survey could be different.
A total of 117 individuals joined our our survey. As the confusion matrix can be seen in Table \ref{table:confusion-matrix}, our synthetics attain a recall rate as low as $64.2\%$, which is close to the ideal case of $50\%$.

{\renewcommand{\arraystretch}{1.3}
\begin{table}[t]
\begin{center}
\begin{tabular}{l|l|c|c|c}
\multicolumn{2}{c}{}&\multicolumn{2}{c}{Prediction}&\\
\cline{3-4}
\multicolumn{2}{c|}{}&Positive&Negative&\multicolumn{1}{c}{Total}\\
\cline{2-4}
\multirow{2}{*}{Ground truth}& Positive & $1057$ & $589$ & $1646$\\
\cline{2-4}
& Negative & $698$ & $1166$ & $1864$\\
\cline{2-4}
\multicolumn{1}{c}{} & \multicolumn{1}{c}{Total} & \multicolumn{1}{c}{$1755$} & \multicolumn{    1}{c}{$1755$} & \multicolumn{1}{c}{~}\\
\end{tabular}
\\
\end{center}
\caption{Confusion matrix of the survey results, where \textit{Positive} and \textit{Negative} denote the synthesized and realistic snowy images, respectively.}
\label{table:confusion-matrix}
\end{table}
}

\section{Experiments} \label{experiment}
The dataset, source code, trained weights will be release at \href{https://goo.gl/BrRc3U}{project website}\footnote{project website: https://goo.gl/BrRc3U}.

\subsection{Implementation Details}

\textbf{Descriptor.} 
We set the kernel size and stride of all polling operations in Inception-v4 \cite{DBLP:inceptionv4} to $3 \times 3$ and $1 \times 1$, respectively, to maintain the spatial information.  
Also, $\gamma$ as defined in Eq. (\ref{eq:DP}) is set at 4. 

\textbf{Recovery submodule.} 
We set $\beta$ as defined in Eqs. (\ref{eq:pyramid_maxout}) and (\ref{eq:r}) at 4 so that the kernel sizes range from $\text{conv}_{1 \times 1}$ to $\text{conv}_{7 \times 7}$. 
We implement the convolution kernels of sizes $\text{conv}_{5 \times 5}$ and $\text{conv}_{7 \times 7}$ in both the pyramid maxout and pyramid sum with vectors of sizes $\text{conv}_{1 \times 5}$ and $\text{conv}_{1 \times 7}$, respectively, to increase speed without reducing accuracy \cite{DBLP:inceptionv4}. 
We use PReLU \cite{he2015delving} as the activation function on the outputs of SE and AE.

\begin{figure}[t]
	\centering
    \includegraphics[width=0.8\linewidth]{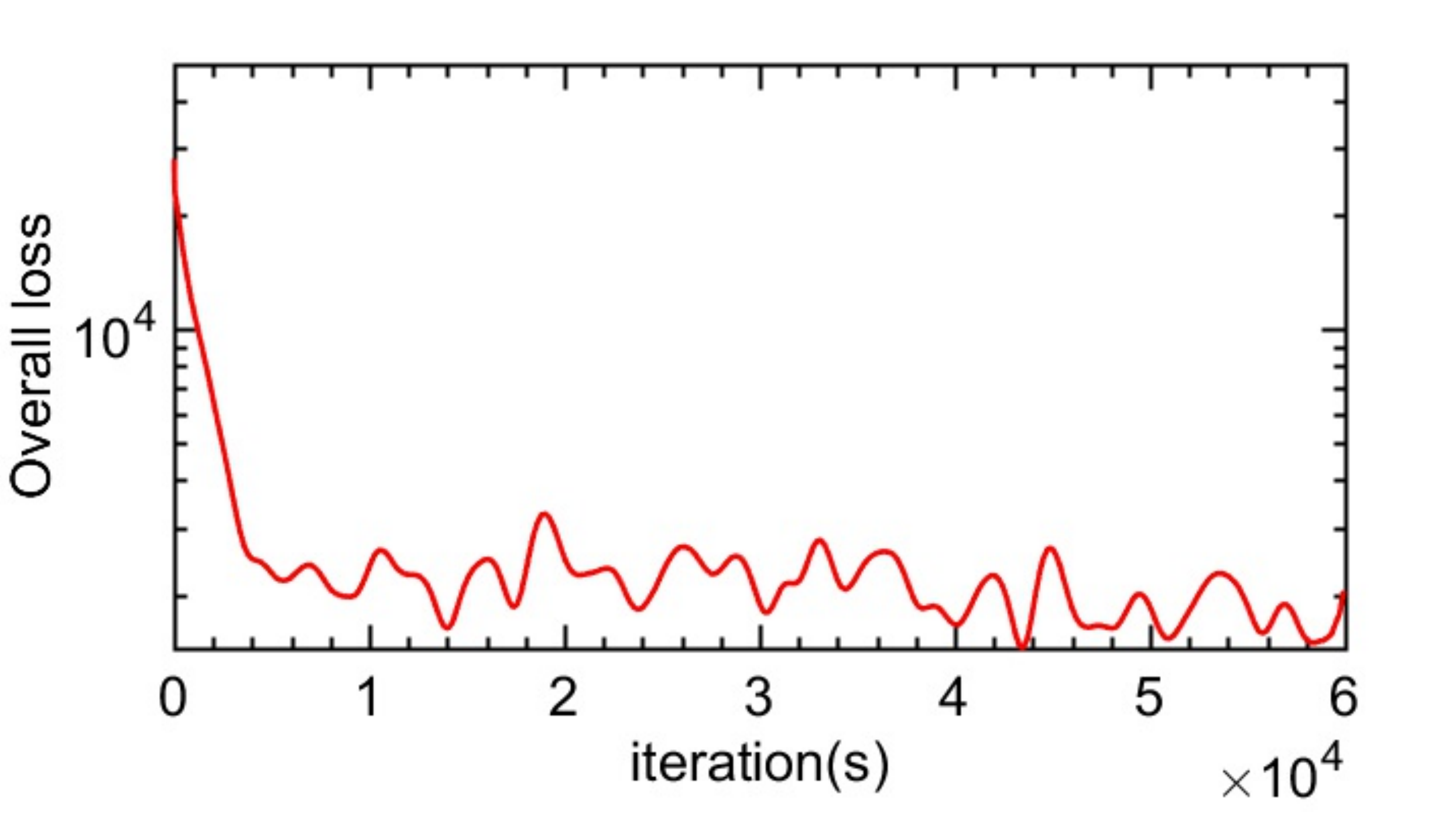}
  	\caption{Training convergence curve of $\mathcal{L}_{overall}$ as defined in Eq. (\ref{eq:overall_loss}).}
    \label{fig:loss_curve}
\end{figure}

\textbf{Training details.} 
The size of the training batch is set at 5, and we randomly crop a patch of size $64 \times 64$ from the samples in Snow100K's training set as the training sample. 
The learning rate is set at $3e^{-5}$ with the Adam Optimizer \cite{kingma2014adam}, and the weightings of our model are initialized by Xavier Initialization \cite{glorot2010understanding}.
Fig. \ref{fig:loss_curve} illustrates the training convergence of the loss defined in Eq. (\ref{eq:overall_loss}) over iterations, where $\lambda_{\mathbf{\hat{z}}}=3$ and $\lambda_{\mathbf{w}}=5e^{-4}$ are set in our work. 

\subsection{Ablation Studies}

Tables \ref{table:ablation_descriptor}-\ref{table:ablation_activation} illustrate the influences of different factors in the proposed DesnowNet (abbr. DSN). 
Two widely used metrics, PSNR and SSIM, are adopted to evaluate the snow removal quality regarding signal and structure similarities. 
Specifically, we evaluate the difference between the snow-free output $\mathbf{\hat{y}}$ and the corresponding ground truth $\mathbf{y}$ via these two metrics. 
In ablation studies, we randomly selected 2k samples from within each of the test subsets of Snow100K-L, Snow100K-M, and Snow100K-S for evaluation, so that there are a total of 6k samples for deriving averages.
Notably, only the evaluated terms are modified in the following experiments, and the remainder of the proposed DesnoNet is kept identical to the parameters introduced in Section \ref{proposed method}. 

\begin{figure*}[t]
	\begin{center}
      \begin{subfigure}{.2\textwidth}
        \centering
        \includegraphics[width=.97\linewidth]{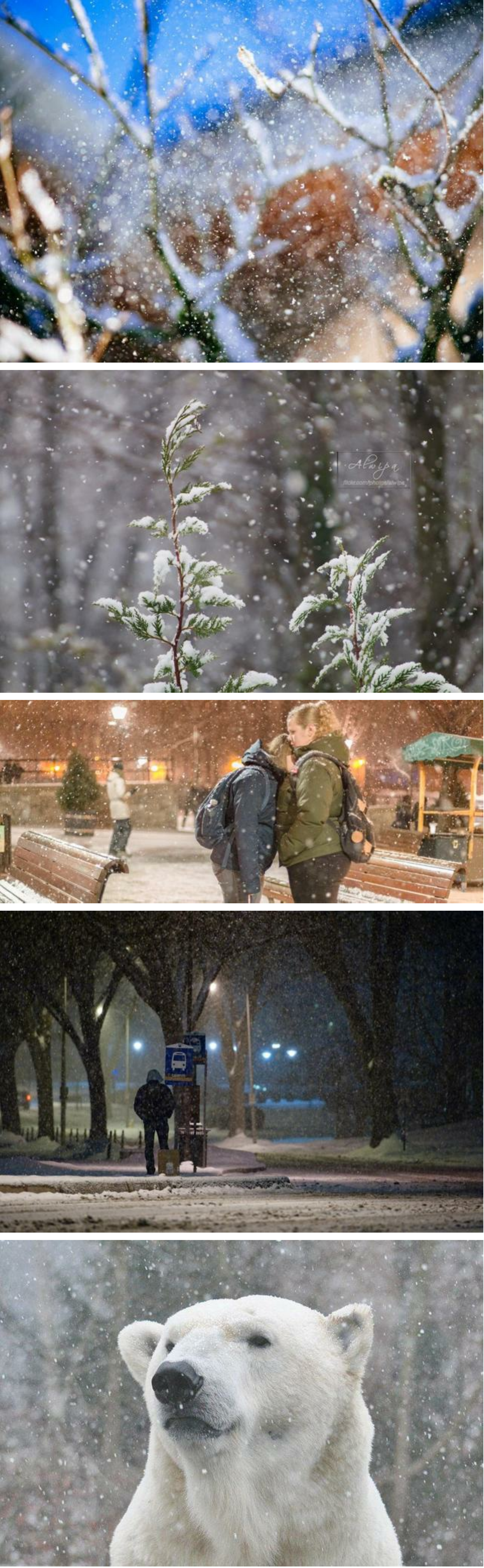}
        \caption{Realistic snowy image ($\mathbf{x}$)}
      \end{subfigure}%
      \begin{subfigure}{.2\textwidth}
        \centering
        \includegraphics[width=.97\linewidth]{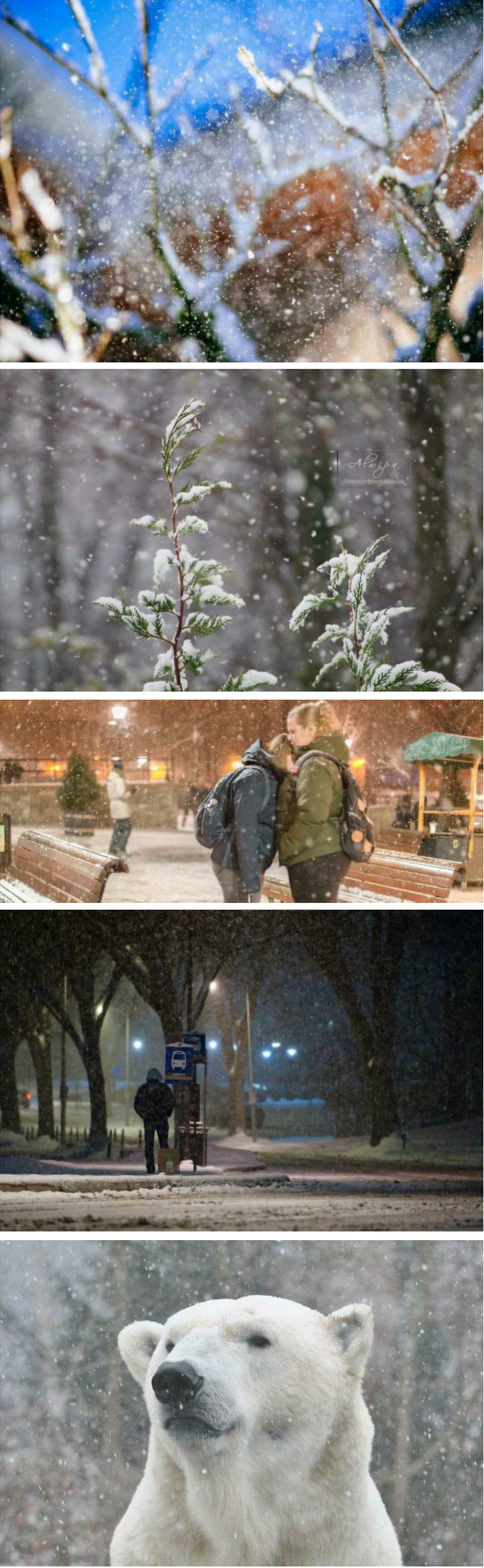}
        \caption{DerainNet}
      \end{subfigure}%
      \begin{subfigure}{.2\textwidth}
        \centering
        \includegraphics[width=.97\linewidth]{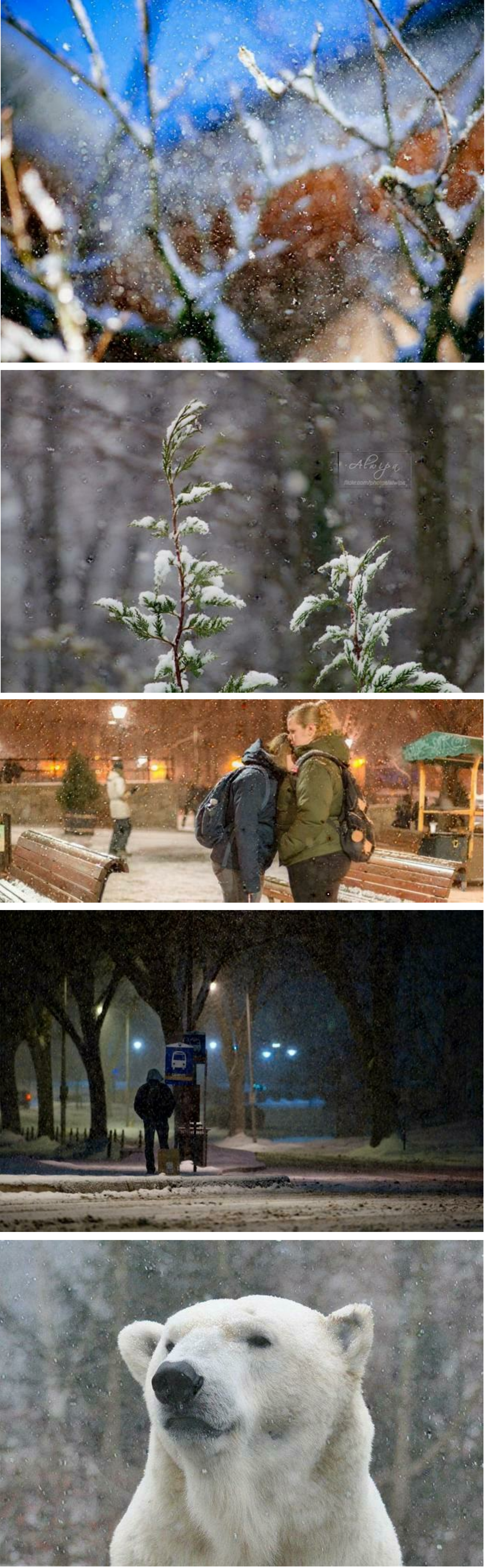}
        \caption{DehazeNet}
      \end{subfigure}%
      \begin{subfigure}{.2\textwidth}
        \centering
        \includegraphics[width=.97\linewidth]{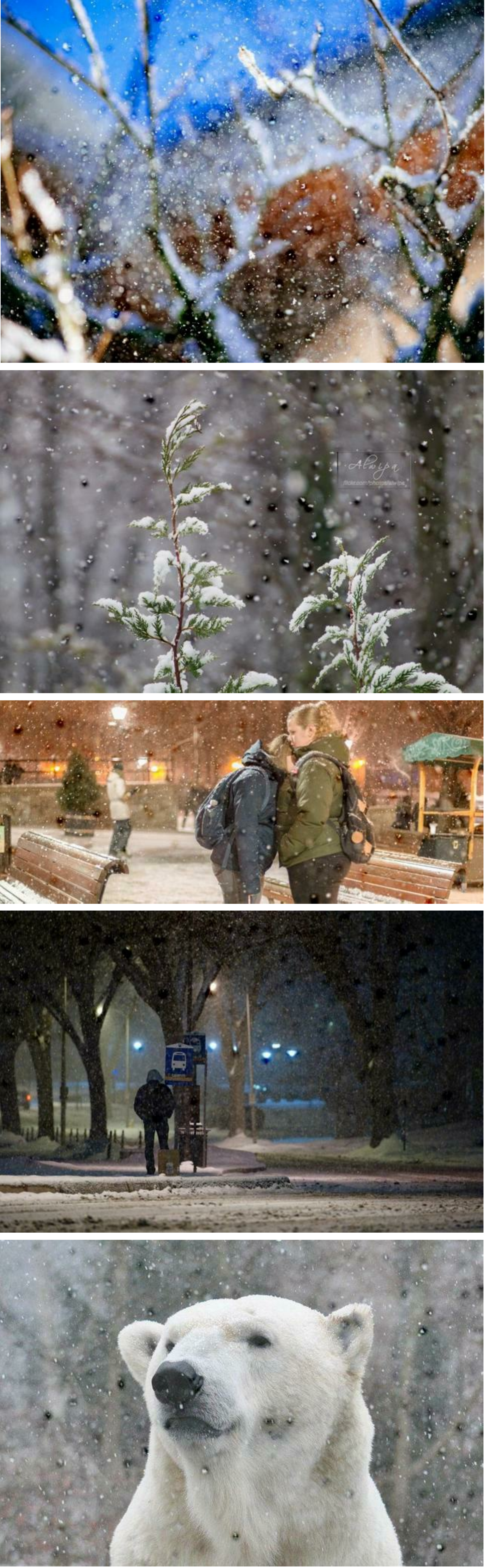}
        \caption{DeepLab}
      \end{subfigure}%
      \begin{subfigure}{.2\textwidth}
        \centering
        \includegraphics[width=.97\linewidth]{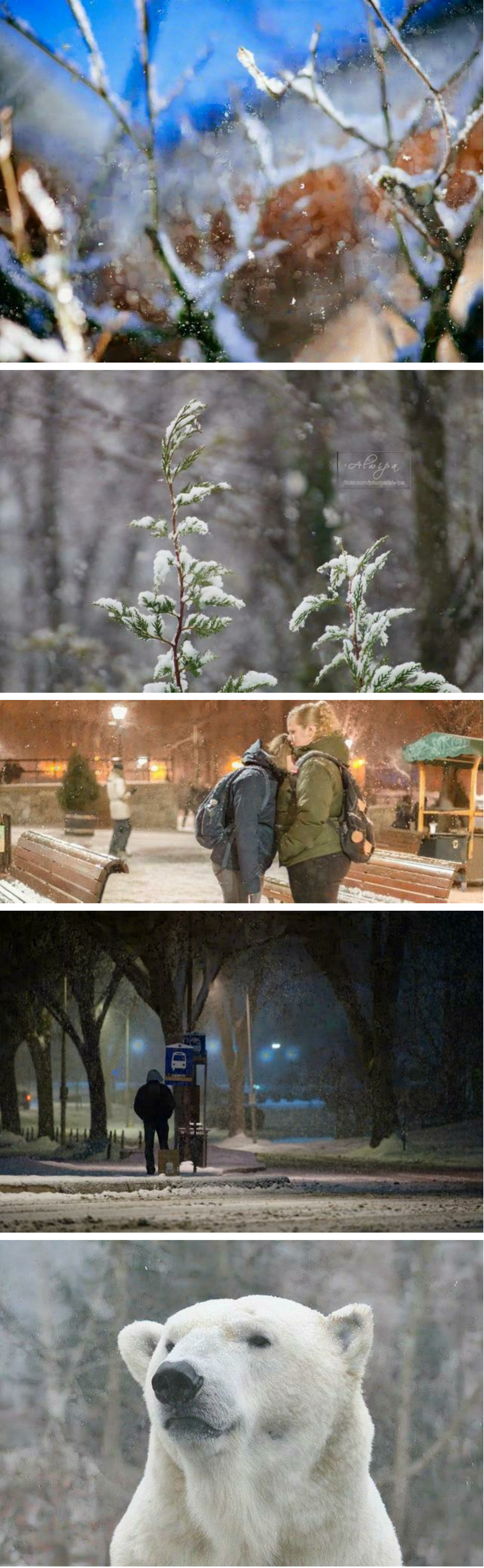}
        \caption{Ours}
      \end{subfigure}%
  	\caption{ Estimated snow-free results $\mathbf{\hat{y}}$ of state-of-the-art methods for realistic snowy images.}
    \label{fig:real_desnow}
	\end{center}
\end{figure*}

\textbf{Descriptor.} 
Table \ref{table:ablation_descriptor} shows the performances achieved using different backbones for descriptors $D_t$ and $D_r$. 
As can be seen, the concatenation operation in the proposed dilation pyramid (DP) defined in Eq. (\ref{eq:DP}) can nicely aggregate the feature maps with multi-scale dilated convolutions and reaches the highest PSNR and SSIM.
{\renewcommand{\arraystretch}{1.3}
\begin{table}[t]
\begin{center}
\begin{tabular}{|l|c|c|c|c|c|c|}
\hline
Metric &  PSNR & SSIM \\
\hline
\hline
DSN w/ Inception-v4 \cite{DBLP:inceptionv4} & 27.898 & 0.8937  \\ %
DSN w/ I.-v4 + ASPP \cite{deeplab}          & 28.7691 & 0.9181 \\ %
DSN w/ I.-v4 + DP                           & 30.1741 & 0.9303  \\
\hline
\end{tabular}
\\
\end{center}
\caption{Comparison of different descriptors in DesnowNet.}
\label{table:ablation_descriptor}
\end{table}
}

{\renewcommand{\arraystretch}{1.3}
\begin{table}[t]
\begin{center}
\begin{tabular}{|l|c|c|c|c|c|c|}
\hline
Metric &  PSNR & SSIM \\
\hline
\hline
DSN w/ $\text{conv}_{3\times3}$    & 28.2275 & 0.9155  \\
DSN w/ pyramid sum     & 28.5529 & 0.9232 \\
DSN w/ pyramid maxout  & 30.1741 & 0.9303  \\
\hline
\end{tabular}
\\
\end{center}
\caption{Comparison of different networks on SE and AE.}
\label{table:ablation_pyramid-maxout}
\end{table}
}

\textbf{Pyramid maxout.} 
Table \ref{table:ablation_pyramid-maxout} shows the performances of different networks for both SE and AE. 
To evaluate the influence without the pyramid maxout, we implement a basic $\text{conv}_{3\times3}$ layer to convert the dimensions of the feature map to meet that of the next layer. 
In addition, we change the pyramid maxout to the pyramid sum as defined in Eq. (\ref{eq:r}) in order to observe the influence. 
As can be seen, pyramid sum reaches a similar performance to that of $\text{conv}_{3\times3}$ because it either cannot account for the variation in snow particle size and shape or lacks sufficient sensitivity for this task.  
However, this property is not necessary for predicting the snow mask $\mathbf{z}$ and chromatic aberration map $\mathbf{a}$ since these two feature maps are more concerned with accuracy rather than visual quality, which the purpose of the pyramid sum discussed at the end of Section \ref{sec:recovery-module}. 
In contrast, the pyramid maxout yields robust prediction in this situation, with the highest PSNR and SSIM. 

{\renewcommand{\arraystretch}{1.3}
\begin{table}[t]
\begin{center}
\begin{tabular}{|l|c|c|}
\hline
Metric & PSNR & SSIM \\
\hline
\hline
DSN w/o TR     & 27.5136 & 0.8911  \\   %
DSN w/o RG     & 28.0149 & 0.8974  \\    %
DSN w/o AE     & 29.086 & 0.9088  \\ %
DSN            & 30.1741 & 0.9303  \\
\hline
\end{tabular}
\\
\end{center}
\caption{Comparison of different architectures of DesnowNet.}
\label{table:ablation_recovery}
\end{table}
}

\begin{figure*}[t]
      \begin{subfigure}{.23\textwidth}
        \centering
        \includegraphics[width=.97\linewidth]{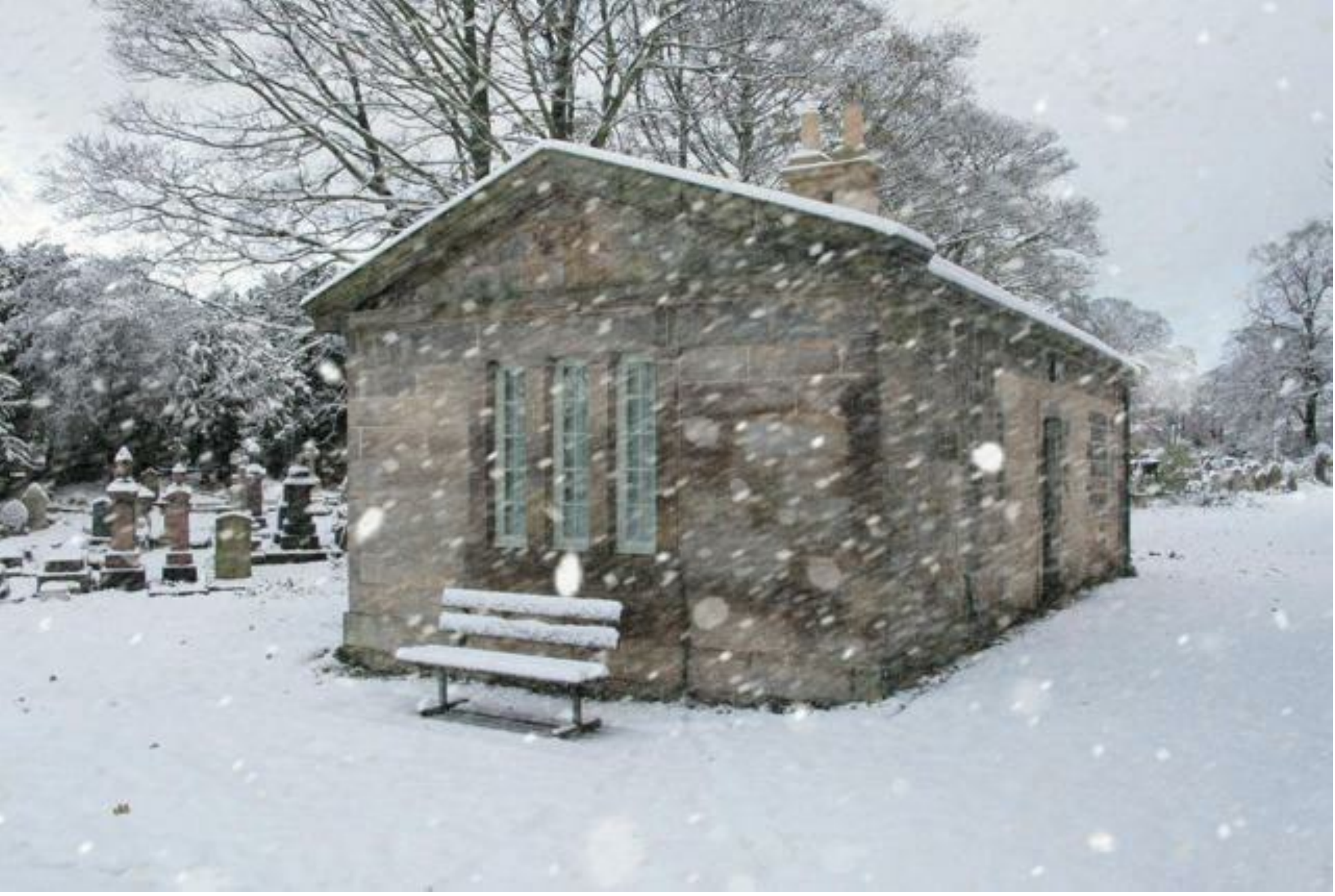}
        \caption{Synthesized snowy image ($\mathbf{x}$)}
      \end{subfigure}%
      \begin{subfigure}{.23\textwidth}
        \centering
        \includegraphics[width=0.97\linewidth]{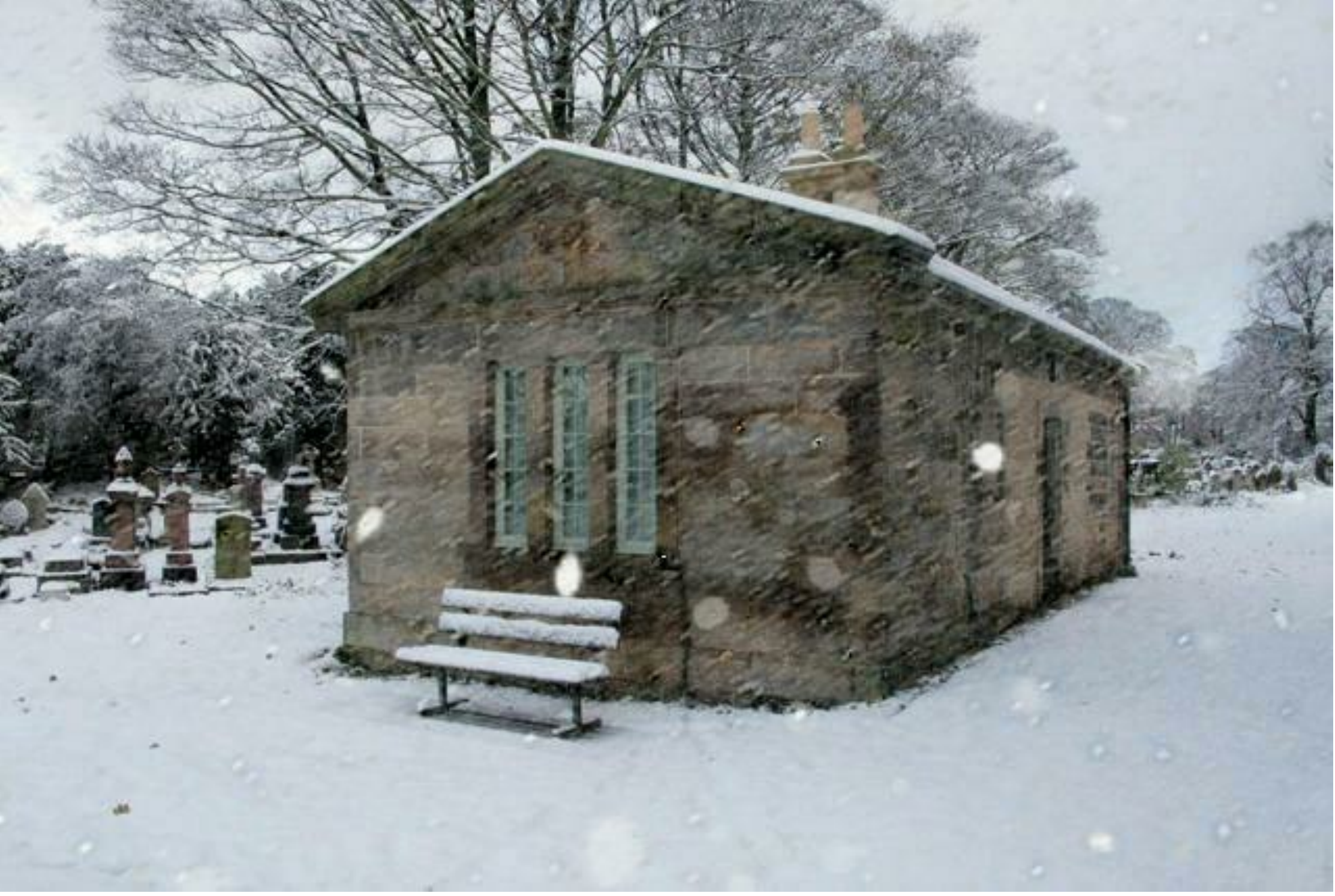}
        \caption{DehazeNet ($\mathbf{\hat{y}}$)}
      \end{subfigure}%
      \begin{subfigure}{.23\textwidth}
        \centering
        \includegraphics[width=0.97\linewidth]{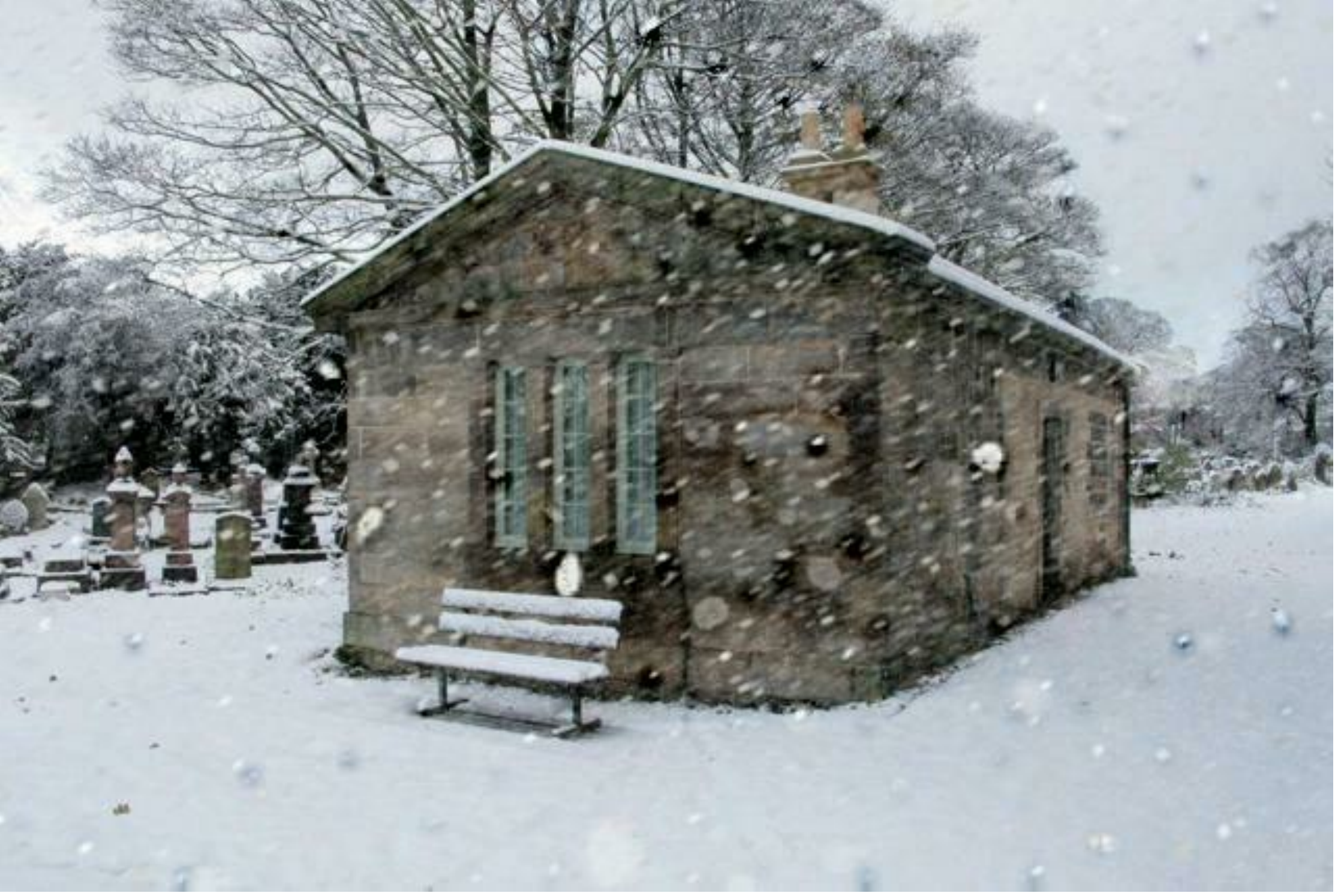}
        \caption{DeepLab ($\mathbf{\hat{y}}$)}
      \end{subfigure}%
      \begin{subfigure}{.23\textwidth}
        \centering
        \includegraphics[width=0.97\linewidth]{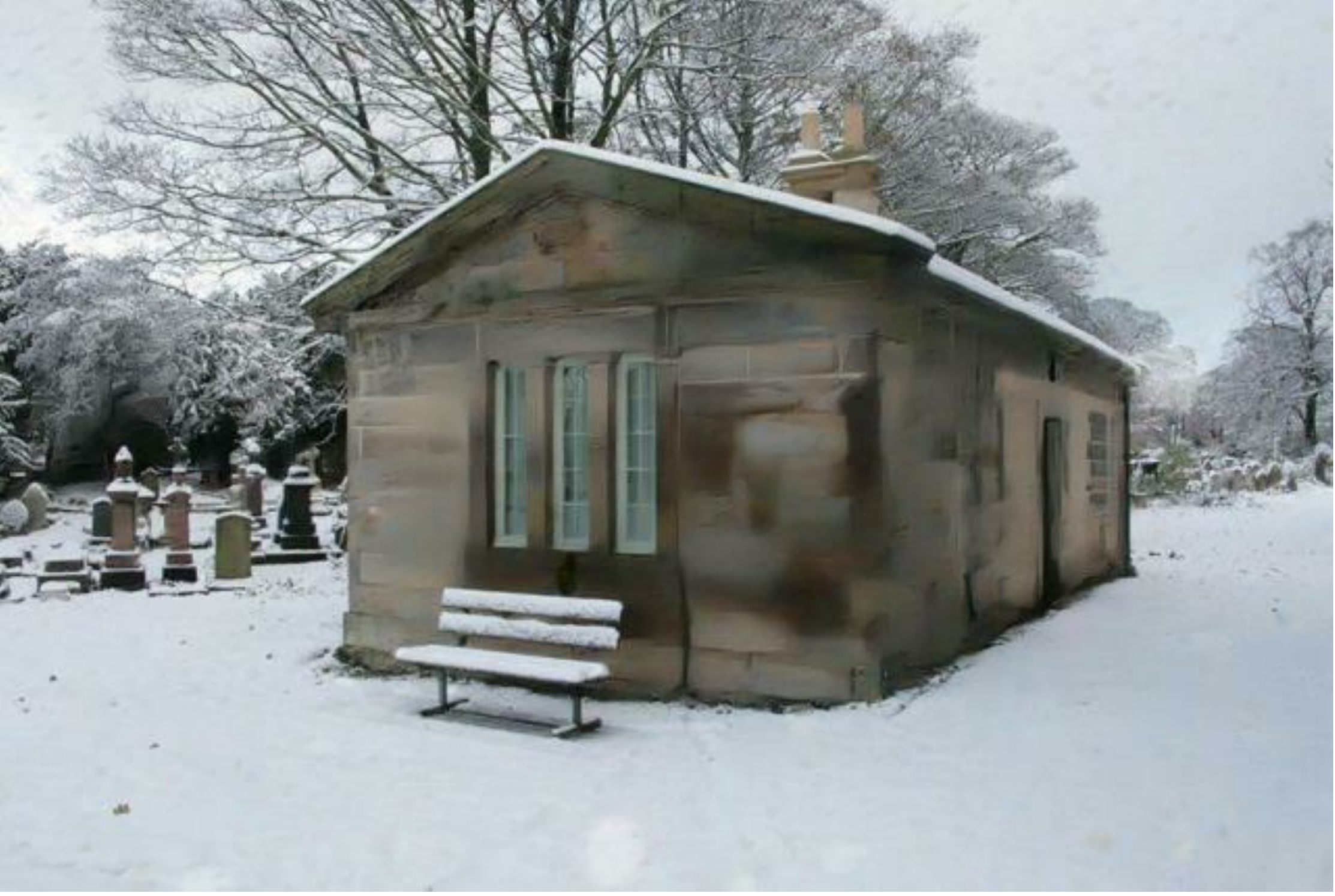}
        \caption{Ours ($\mathbf{\hat{y}}$)}
      \end{subfigure}  \\%
      
      \begin{subfigure}{.23\textwidth}
        \centering
        \includegraphics[width=.97\linewidth]{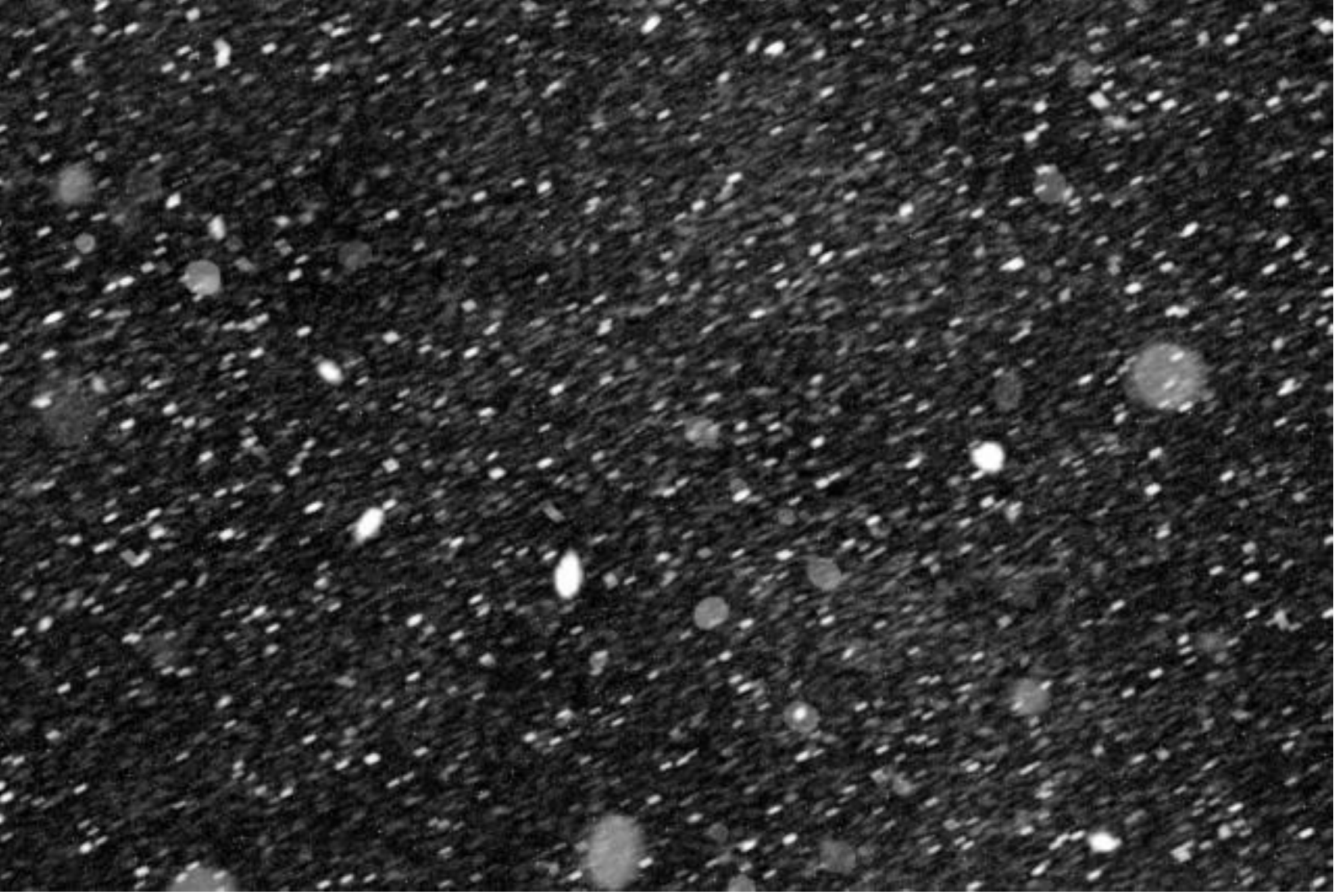}
        \caption{Synthesized snow mask ($\mathbf{z}$)}
      \end{subfigure}%
      \begin{subfigure}{.23\textwidth}
        \centering
        \includegraphics[width=0.97\linewidth]{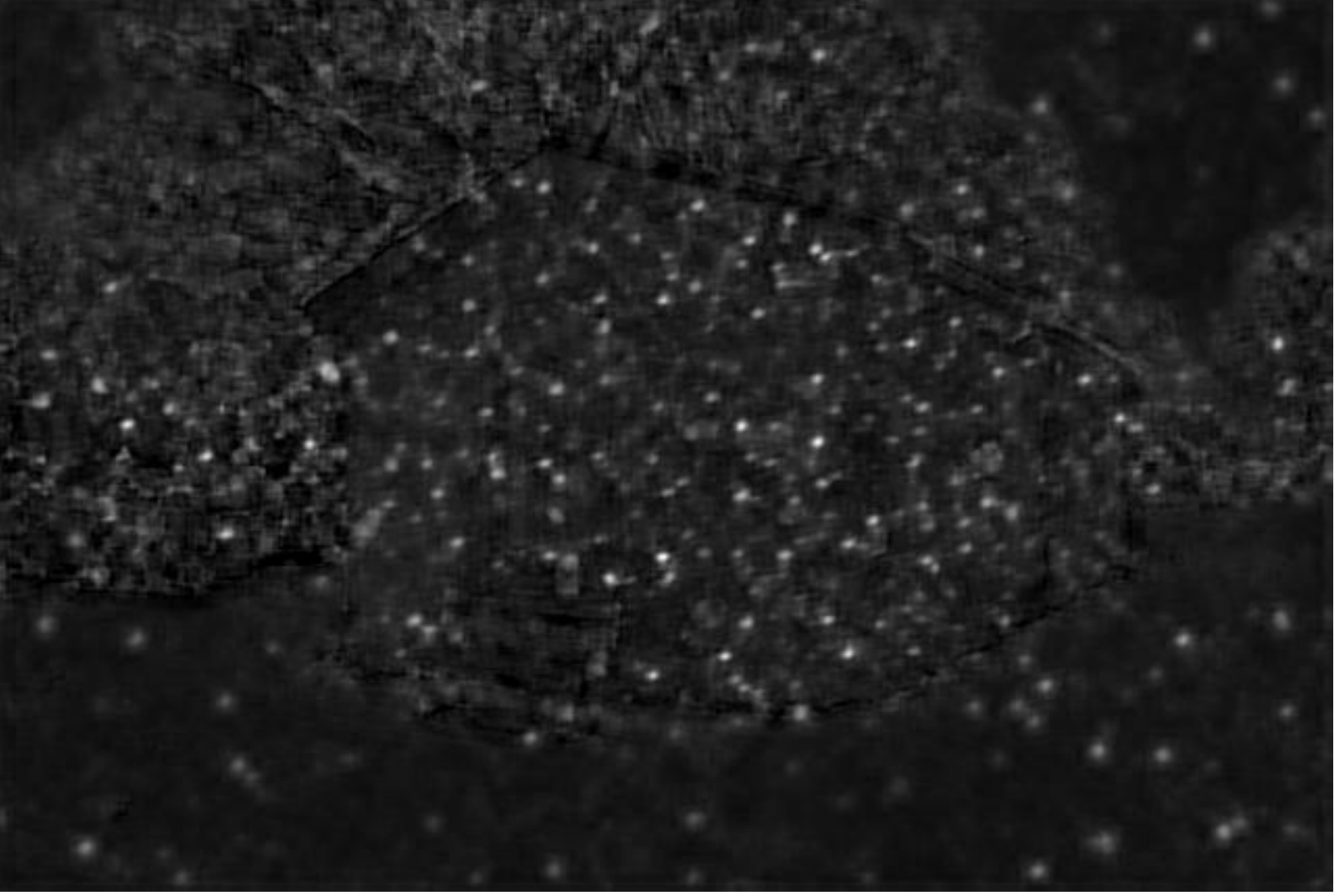}
        \caption{DehazeNet ($\mathbf{\hat{z}}$)}
      \end{subfigure}%
      \begin{subfigure}{.23\textwidth}
        \centering
        \includegraphics[width=0.97\linewidth]{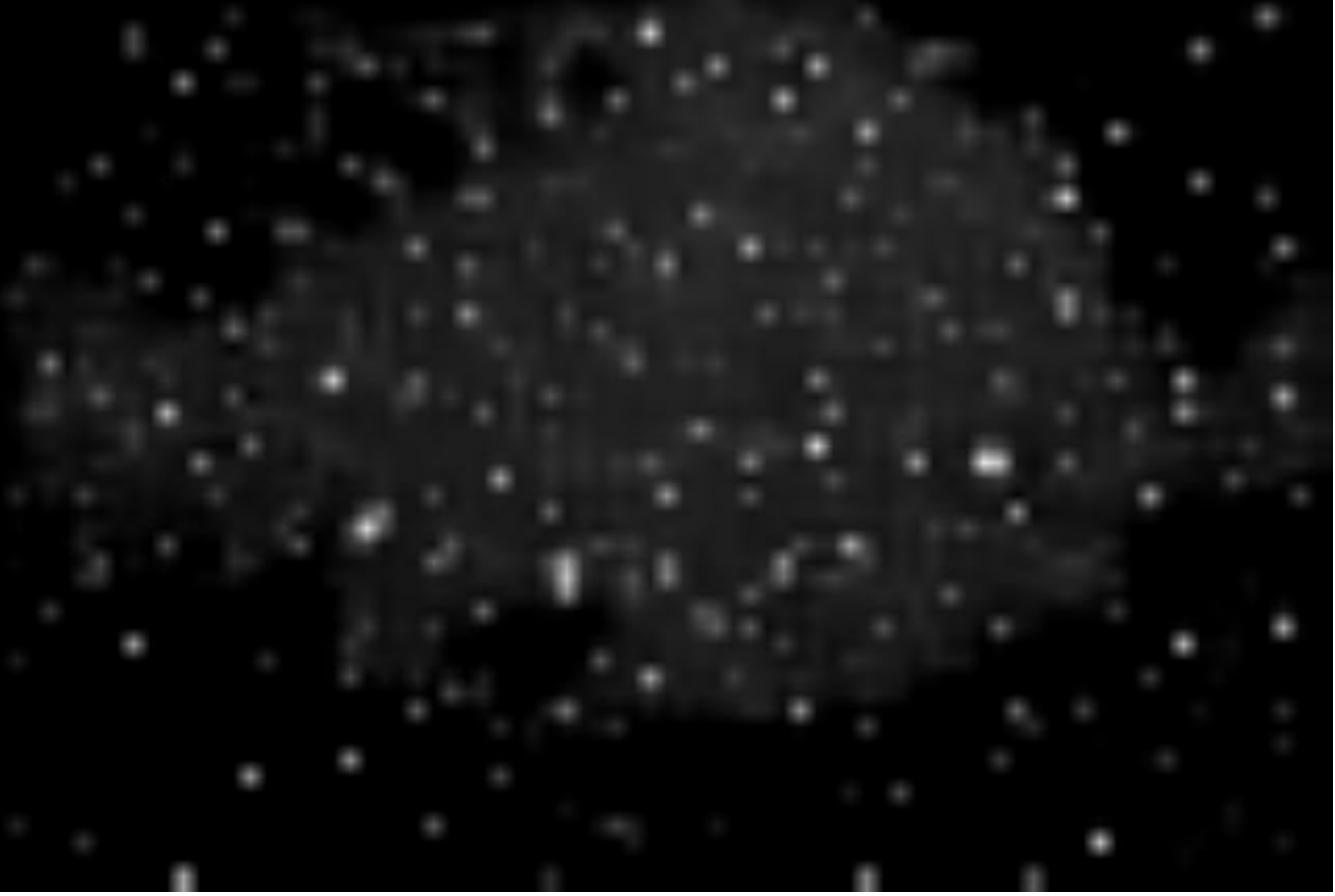}
        \caption{DeepLab ($\mathbf{\hat{z}}$)}
      \end{subfigure}%
      \begin{subfigure}{.23\textwidth}
        \centering
        \includegraphics[width=0.97\linewidth]{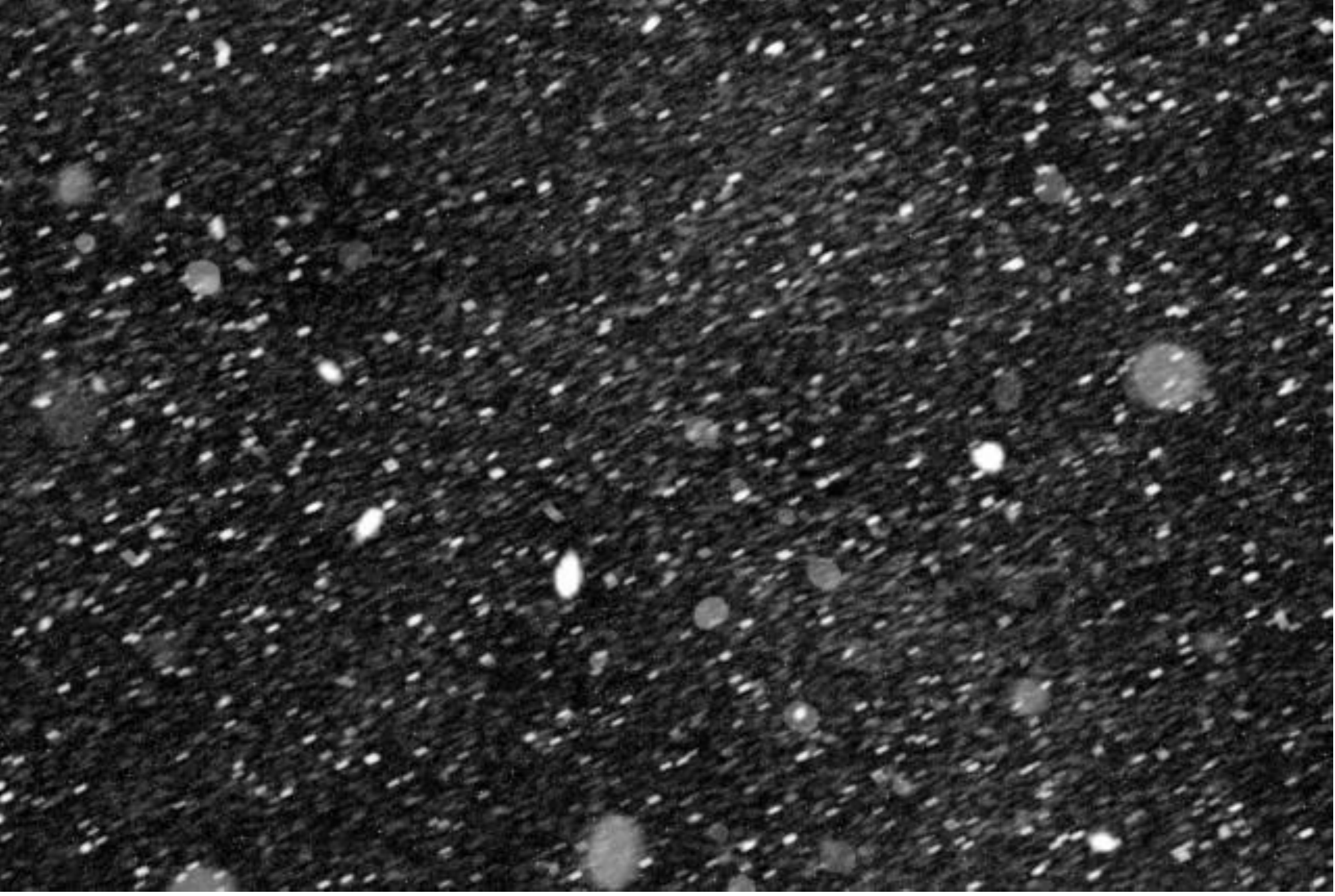}
        \caption{Ours ($\mathbf{\hat{z}}$)}
      \end{subfigure}
      
  	\caption{Snow removal results of the proposed DesnowNet and other state-of-the-art methods.}
	\label{fig:syn_z_compare}
\end{figure*}

\textbf{Recovery module.} 
Certain components of the recovery module have been removed in order to ascertain their potential benefits. Results are shown in Table \ref{table:ablation_recovery}.
By removing the TR module, only the RG module is available for predicting the estimated snow-free result $\mathbf{\hat{y}}$ without the prior of estimated snow mask $\mathbf{\hat{z}}$ and chromatic aberration map $\mathbf{a}$. Results show that this has a large impact on performance. 
On the other hand, while removing AE from DSN results in lower reduction of accuracy in contrast to the removal of other components, it still contributes 1.09 dB at PSNR.

\textbf{Pyramid loss function.}
Table \ref{table:ablation_loss} shows the results of different loss functions, which clearly indicate that the pyramid $l$2-norm as defined in Eq. (\ref{eq:pyramid_loss}) is superior to the $l$2-norm that utilizes only single scale of the feature. 

\textbf{Activation function.}
We also changed the activation function of the outputs of SE and AE to ascertain their influence. 
Table \ref{table:ablation_activation} exhibits the results, which show that the PReLU \cite{he2015delving} achieves significantly better estimation accuracy. 

{\renewcommand{\arraystretch}{1.3}
\begin{table}[t]
\begin{center}
\begin{tabular}{|l|c|c|}
\hline
Metric & PSNR & SSIM \\
\hline
\hline
DSN w/ $l$2-norm & 29.223 & 0.9127 \\
DSN w/ pyramid $l$2-norm  & 30.1741 & 0.9303  \\
\hline
\end{tabular}
\\
\end{center}
\caption{Comparison of different loss functions in DesnowNet.}
\label{table:ablation_loss}
\end{table}
}

{\renewcommand{\arraystretch}{1.3}
\begin{table}[t]
\begin{center}
\begin{tabular}{|l|c|c|}
\hline
Metric & PSNR & SSIM \\
\hline
\hline
Sigmoid &  27.8894 & 0.8985  \\
PReLU \cite{he2015delving} & 30.1741 & 0.9303  \\
\hline
\end{tabular}
\\
\end{center}
\caption{Comparison of different activation functions on the outputs of SE and AE.}
\label{table:ablation_activation}
\end{table}
}

{\renewcommand{\arraystretch}{1.3}
\begin{table*}[t] 
\begin{center}
\begin{tabular}{c|c|c|c|c|c|c|c|c|c|}
\cline{1-9}
\multicolumn{1}{|c|}{Subset} & \multicolumn{2}{c|}{Snow100K-S} & \multicolumn{2}{c|}{Snow100K-M} & \multicolumn{2}{c|}{Snow100K-L}  & \multicolumn{2}{c|}{Overall} \\ \cline{1-9} 
\multicolumn{1}{|c|}{Metric} & PSNR & SSIM & PSNR & SSIM & PSNR & SSIM & PSNR & SSIM \\ \hline \hline
\multicolumn{1}{|c|}{Synthesized data }              & 25.1026 & 0.8627 & 22.8238 & 0.838 & 18.6777 & 0.7332 & 22.1876 & 0.8109 \\ \hline
\multicolumn{1}{|c|}{DerainNet \cite{derain2017} } & 25.7457 & 0.8699 & 23.3669 & 0.8453 & 19.1831 & 0.7495 & 22.7652 & 0.8216 \\ \hline
\multicolumn{1}{|c|}{DehazeNet \cite{dehaze2016} } & 24.9605 & 0.8832 & 24.1646 & 0.8666 & 22.6175 & 0.7975 & 23.9091 & 0.8488 \\ \hline
\multicolumn{1}{|c|}{DeepLab \cite{deeplab} } & 25.9472 & 0.8783 & 24.3677 & 0.8572 & 21.2931 & 0.7747 & 23.8693 & 0.8367 \\ \hline
\multicolumn{1}{|c|}{Ours }                   & 32.3331 & 0.95 & 30.8682 & 0.9409 & 27.1699 & 0.8983 & 30.1121 & 0.9296 \\ \hline
\end{tabular}
\\
\end{center}
\caption{Respective performances of the different state-of-the-art methods evaluated via Snow100K's test set; the results shown in the \textit{Synthesized data} row show the similarities between the synthesized snowy image $\mathbf{x}$ and the snow-free ground truth $\mathbf{y}$, and the \textit{Overall} column presents the averages over the entire test set.}
\label{table:state-of-the-art}
\end{table*}
}

\subsection{Comparison}

Two learning-based atmospheric particle removal approaches, DerainNet \cite{derain2017} and DehazeNet \cite{dehaze2016}, and one semantic segmentation method, DeepLab \cite{deeplab}, were considered in the comparison of estimated snow-free results ($\mathbf{\hat{y}}$) due to their effective generalization capabilities and the competitive visual qualities in their focuses.

Due to a lack of source codes, we re-implemented DerainNet, DehazeNet, and DeepLab with their default parameters to achieve the best respective performances. 
For DerainNet, the detailed part of the snow-free image $\mathbf{y}$ was used as the ground truth for training.
For DehazeNet, we adopted the snow mask $\mathbf{z}$ as its ground truth. 
Also, we use Eq. (\ref{eq:recovery}) with their estimated snow mask $\mathbf{\hat{z}}$ to directly recovery $\mathbf{\hat{y}}$ without our RG module to match their network design. 
We implemented the DeepLab-lfov version for our comparison, and used the snow mask $\mathbf{z}$ as the ground truth for training. 
As for the implementation of DehazeNet, we used Eq. (\ref{eq:recovery}) with DeepLab's estimated $\mathbf{\hat{z}}$ to generate the snow-free estimate $\mathbf{\hat{y}}$.

Table \ref{table:state-of-the-art} presents the quantitative results obtained via three test subsets of the Snow100K dataset. It shows that our DesnowNet outperformed other learning-based methods, with DehazeNet achieving the second-best accuracy rates.
Fig. \ref{fig:real_desnow} shows the results of the estimated snow-free outputs $\mathbf{\hat{y}}$.  
Among these, DerainNet removes almost nothing from the snowy images because of its strong assumption for spatial frequency. 
DehazeNet is effective at removing many translucent and opaque snow particles from snowy images. However, it introduces apparent artifacts in the resultant images.
Results obtained by the DeepLab network exhibit even stronger artifacts than DehazeNet, and even introduces black spots due to the opaque snow particles. 
Conversely, the results show that our method introduced the fewest artifacts in its estimated snow-free outputs $\mathbf{\hat{y}}$, and that it achieves the most visually appealing results. 

An additional experiment was conducted to determine the accuracy of the estimated snow mask $\mathbf{\hat{z}}$. 
Because of the lack of intermediate probability estimation in DerainNet, we excluded it from this comparison.
Table \ref{table:corruption_comparison} presents the average similarity scores between the snow mask ground truth $\mathbf{z}$ and the estimated $\mathbf{\hat{z}}$. 
As can be observed, our method yields a superior similarity score from both PSNR and SSIM metrics.
Of these, the large SSIM superiority particularly contrasts with other and is primarily due to our capability of interpreting variation in the spatial frequency, trajectory, translucency, and the particle size and shape of snows.
The qualitative comparison shown in Fig. \ref{fig:syn_z_compare} validates this superiority. 
In this experiment, DehazeNet was able to interpret only the fine-grained snow particles as shown in Fig. \ref{fig:syn_z_compare}(f). 
While DeepLab can describe snow particles of different sizes as shown in Fig. \ref{fig:syn_z_compare}(g), its down-sampling design loses much spatial information, resulting in its failure to identify small-size snow particles.
Fig. \ref{fig:syn_z_compare}(h) presents the results of our method, which clearly shows that it successfully interpreted all the variations of snow particles to yield an accurate snow-free output, as illustrated in Fig. \ref{fig:syn_z_compare}(d).

{\renewcommand{\arraystretch}{1.3}
\begin{table}[t] 
\begin{center}
\begin{tabular}{|l|c|c|}
\hline
Metric & PSNR & SSIM \\
\hline\hline
DehazeNet \cite{dehaze2016} & 19.622 & 0.3271 \\
DeepLab   \cite{deeplab} & 18.8679 & 0.2942 \\
Ours  & 22.0054 & 0.5662 \\
\hline
\end{tabular}
\end{center}
\caption{Accuracy comparison of the estimated snow mask $\mathbf{\hat{z}}$ on Snow100K's test set.}
\label{table:corruption_comparison}
\end{table}
}

\subsection{Failure Cases} \label{sec:failure_cases}

Some situations occurred in which our method generated speckle-like artifacts on realistic images, as exhibited in Fig. 11. 
These were primarily caused by confusing and blurry backgrounds (non-covered areas) that made it difficult to recover obscured regions. However, in most cases this does not detract from the essential purpose of our work – distinguishing image details such as individuals and their behavior obscured by falling snow.

\begin{figure}[t]
	\begin{center}
        \centering
        \includegraphics[width=.97\linewidth]{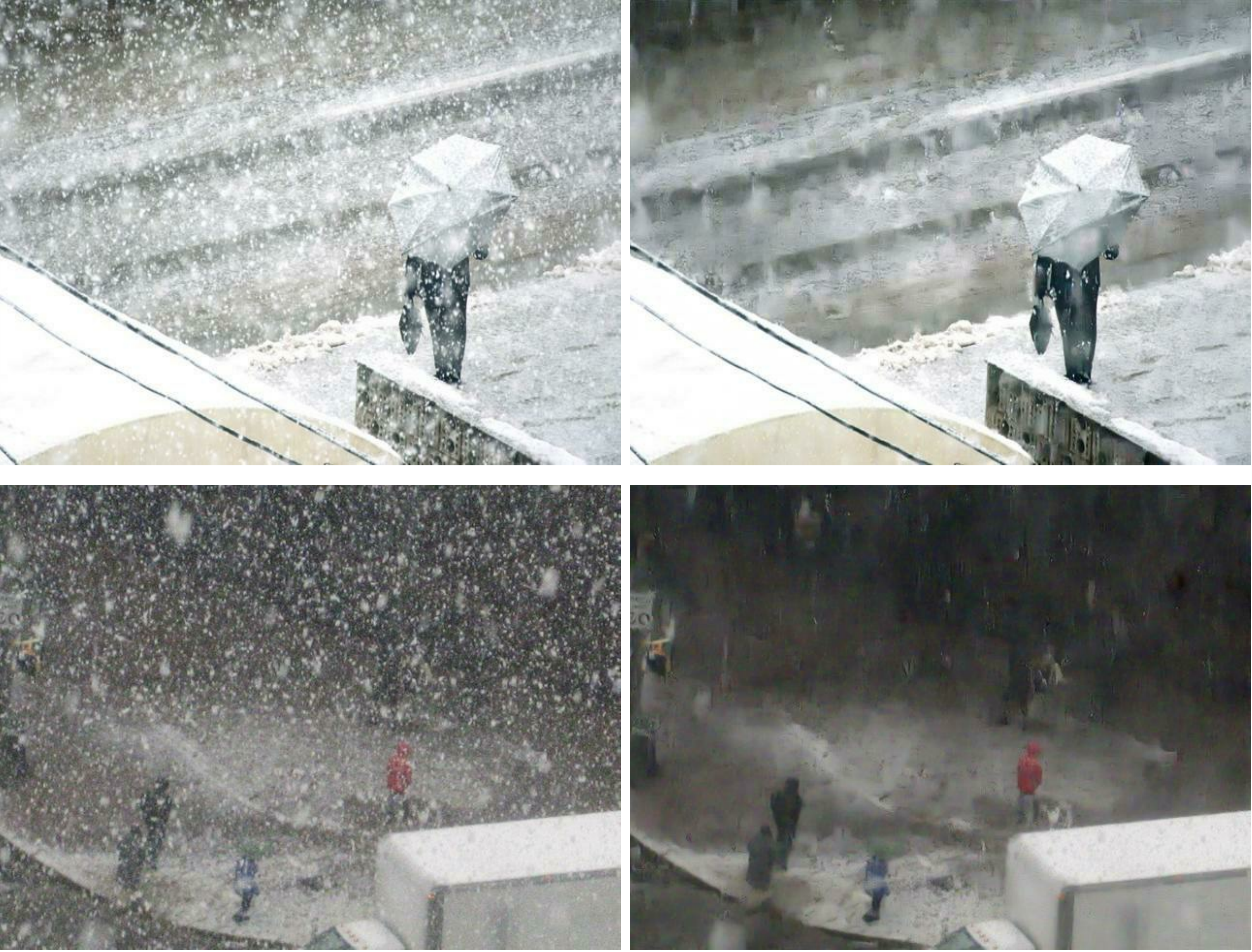}
  	\caption{Failure cases $\mathbf{\hat{y}}$ (right) of the proposed DesnowNet, where the left images show the corresponding realistic snowy images $\mathbf{x}$.}
    \label{fig:real_failure}
	\end{center}
\end{figure}

\section{Conclusions} \label{conclusion}

We have presented the first learning-based snow removal method for the removal of snow particles from single image. Also, we have demonstrated that the baseline of a well-known semantic segmentation method and other state-of-the-art atmospheric particle removal approaches, are unable to adapt to the challenging snow removal task. 
We observe the multistage network design with translucency recovery (TR) and residual generation (RG) modules successfully recovers image details obscured by opaque snow particles as well as compensates for the potential artifacts. 
In addition, the results presented in Table \ref{table:ablation_recovery} demonstrate that the chromatic aberration map, which interprets subtle color inconsistencies of snow in three color channels as shown in Fig. \ref{fig:results_hists}(a), is the key milestone towards accurate prediction. Moreover, our multi-scale designs endow the network with interpretability that can account for variations of snow particles, as demonstrated in Fig. \ref{fig:syn_z_compare}. 
In certain situations, our method may introduce speckle-like artifacts (discussed in Section \ref{sec:failure_cases}), which could be a possible area of future research in the pursuit of better visual quality.


\bibliographystyle{ACM-Reference-Format}
\bibliography{main}
\end{CJK} 
\end{document}